\documentclass{IEEEtran}
\usepackage{graphicx} 
\usepackage{amsfonts} 
\usepackage{amsmath}
\usepackage{mathrsfs}
\usepackage{xcolor}  
\usepackage{pgfplots}
\usepackage{orcidlink}
\usepackage[ruled,vlined]{algorithm2e}
\usepackage{url}
\usepackage{hyperref}
\pgfplotsset{compat=1.18}
\usepgfplotslibrary{fillbetween}
\usetikzlibrary{pgfplots.groupplots}
\pgfplotsset{compat=newest}
\usepackage[normalem]{ulem}
\usepackage{orcidlink}
\usepackage[english]{babel}
\usepackage[numbers,sort]{natbib}
\usepackage{multibib}
\newcites{supp}{Supplementary References}

\bstctlcite{BSTcontrol}
\begin{document}
\bstctlcite{BSTcontrol}

\title{Are you sure? A Comprehensive and Comprehensible Survey of Uncertainty Quantification in Symbolic Regression}

\author{Julia Reuter~\orcidlink{0000-0002-7023-7965} and Fabricio Olivetti de Franca~\orcidlink{0000-0002-2741-8736}%
\thanks{This research was funded by Funda\c{c}\~{a}o de Amparo \`{a} Pesquisa do Estado de S\~{a}o Paulo (FAPESP) grant number 2025/19085-3 and CNPq 301596/2022-0.
The authors are with Center for Mathematics, Computation and Cognition (CMCC), Federal University of ABC, Brazil (e-mail: \{julia.reuter,folivetti\}@ufabc.edu.br)}}%

\maketitle
\IEEEpeerreviewmaketitle

\begin{abstract}
  Symbolic regression (SR) is a class of methods that systematically explore the space of mathematical functions to discover models that accurately capture the underlying relationships in a dataset.
  Despite recent advances in the field, a lack of support for uncertainty quantification (UQ) limits its adoption in real-world decision processes.
  In regression analysis, UQ provides important information about the model reliability, which can both help to avoid overfitting by accounting for uncertainty in the data, and provide insights for decision-making.
  This survey is the first to clearly address this issue, with the objective of introducing essential UQ concepts and reviewing the current literature on UQ in SR, which can be broadly organized into three research directions: frequentist, Bayesian, and model selection. 
  Despite its importance, UQ in SR is still underexplored, which motivates further research into reliable UQ methods for SR.
\end{abstract}


\section{Introduction}
\IEEEPARstart{S}{ymbolic regression} (SR)~\cite{koza_genetic_1992,kronberger_symbolic_2024} is a class of search algorithms that systematically traverse the space of mathematical functions, referred to as the \textbf{hypothesis space} $\mathcal{H}$, with the objective of finding the best \textbf{hypothesis} $h^* \in \mathcal{H}$ according to predefined objectives, such as a minimum approximation error on the data, while maintaining low complexity.
In this context, a dataset is a set of $N$ points $\mathcal{D} = \{\mathbf{x}_i, y_i\}_{i=1..N}$, where $\mathbf{x}_i \in \mathbb{R}^{n_x}$ are the $n_x$ independent variables and $y_i \in \mathbb{R}$ is the dependent variable. 
Modern SR typically represents a hypothesis as a function $f(\mathbf{x}; \boldsymbol{\theta})$, where $\boldsymbol{\theta}$ are adjustable parameters used to fit the function into the data.
The goal of SR is to identify a function $f(\mathbf{x}; \boldsymbol{\theta}) \approx y$ within $\mathcal{H}$ that best explains the data. 
The best model found by the SR algorithm is denoted by $\hat{h} \in \mathcal{H}$, whereas $h^*$ is the best model in $\mathcal{H}$~\cite{hullermeier_aleatoric_2021}.

In a perfect world, the learning problem would be free of uncertainties. 
However, SR faces many sources of uncertainty, such as the \textbf{aleatoric uncertainty}, which is caused by  measurement imprecision or intrinsic randomness of the data generation process, and cannot be reduced by collecting more data. 
Although irreducible, this uncertainty can be quantified to estimate the expected noise-induced lower bound of the error of $h^*$. 
Another source is the \textbf{epistemic uncertainty}, which refers to a lack of knowledge and can, in principle, be reduced by acquiring additional information or data.
It can be decomposed into \textbf{model uncertainty}, that refers to the risk that the true model $h^*$ is not present in $\mathcal{H}$, and the \textbf{approximation uncertainty}, representing the uncertainty about a hypothesis $h$ being the best possible within $\mathcal{H}$. A key aspect of the approximation uncertainty is the \textbf{parameter uncertainty}, related to the uncertainty about the values of the adjustable parameters.
Its magnitude is not only influenced by the data, but also by the model structure, which shapes the geometry of the parameter likelihood surface: 
flat or ill-conditioned likelihood surfaces lead to larger parameter uncertainty.
While aleatoric and epistemic uncertainty are conceptually different, their strict separation in practice is often difficult, as both depend on the available data and model assumptions~\cite{hullermeier_aleatoric_2021}.

In SR literature, uncertainties are often not explicitly addressed, as the focus of many works is an efficient traversal of the multimodal search space.
However, uncertainties are a relevant topic, as they directly affect the search and practical use of the generated models. 
Therefore, this paper has the objective of clearly introducing the topic of \textbf{uncertainty quantification}~(UQ) in the context of SR, and to provide a structured overview of existing approaches. 
To this end, we survey the current SR literature with respect to UQ and aim to raise awareness for this important aspect of regression analysis within the SR community, potentially promoting advancements in this field.

\section{Uncertainty quantification in regression}

UQ plays important roles in regression analysis, that are also relevant for SR, such as assessing the confidence about the selected models and their predictions.
In real world scenarios, quantifying the epistemic uncertainty using prediction intervals provides information whether the range of likely values is within some safety bounds imposed by the application, as opposed to point predictions. 
For noisy datasets, quantifying aleatoric uncertainty can help to detect potential overfitting whenever a model prediction error is below this value.
The following subsection introduces the idea of likelihood functions and Fisher information as a foundation for the statistical UQ approaches in Sec.~\ref{sec:frequentist} and~\ref{sec:bayesian}.

\subsection{Likelihood function and Fisher information matrix}
\label{sec:likelihood_fisher}

In classical statistics literature, a data-generating process typically follows a parametrized family of probability measures $P_\theta$, with the parameter vector $\boldsymbol{\theta} \in \boldsymbol{\Theta}$.
The density (or mass) function of $P_\theta$ is denoted by $p_\theta(\cdot)$~\cite[Sec.~4.2]{hullermeier_aleatoric_2021}. 
In regression analysis, a response variable is typically modeled as a function of input variables and a set of $d$ (unknown) parameters~$\boldsymbol{\theta} \in \mathbb{R}^d$ with additive noise, i.e., $y_i = f(\mathbf{x}_i;\boldsymbol{\theta}) + \varepsilon_i $.
This induces a conditional distribution $p_{\boldsymbol{\theta}}(y_i \mid \mathbf{x}_i)$, where $\mathbf{x}_i$ is treated as given, and only the distribution of $y_i$ conditional on $\mathbf{x}_i$ is modeled. 
In this setting, the \textbf{likelihood}~\cite[Sec.~2.3]{pawitan_all_2001} is defined as the joint density of $N$ independent observations:
    \begin{align}
        L_N(\boldsymbol{\theta}) =
        \prod_{i=1}^{N}
        p_{\boldsymbol{\theta}}(y_i \mid \mathbf{x}_i).
    \end{align}

The log-transformation of the likelihood function enables a simplified differentiation and numerical stability:
    \begin{align}
        \ell_N(\boldsymbol{\theta})
        &=\sum_{i=1}^{N}
        \log p_{\boldsymbol{\theta}}(y_i \mid \mathbf{x}_i).
        \label{eq:log_likelihood}
    \end{align}

Typically, observations are assumed to be independent and identically distributed (i.i.d.), i.e., all data points are sampled from the same distribution and bear no influence on each other. 
Moreover, the noise is commonly assumed to follow a Gaussian distribution $\varepsilon_i \sim \mathcal{N}(0,\sigma_i^2)$ with variance $\sigma_i^2$.
Another standard assumption is \textbf{homoscedasticity}, stating that the variance in the response variable is constant across all values of the input variables.
In this case, repeated measurements at any $\mathbf{x}_i$ produce Gaussian-distributed responses $y_i$ with a constant $\sigma^2$ independent of $\mathbf{x}_i$~\cite[Sec.~1.3]{bates_nonlinear_1988}.
Under these assumptions, the Gaussian density is given by~\cite[Sec.~2.3]{pawitan_all_2001}
    \begin{align}
        p_{\boldsymbol{\theta}}(y_i \mid \mathbf{x}_i) &= \frac{1}{\sqrt{2\pi\sigma^2}} \exp\!\left(-\frac{1}{2\sigma^2} \bigl(y_i - f(\mathbf{x}_i;\boldsymbol{\theta})\bigr)^2 \right) ,
        \label{eq:p_theta_reg}
    \end{align}
and the corresponding log-likelihood is
    \begin{align}
        \ell_N(\boldsymbol{\theta}) =
        -\frac{1}{2\sigma^2}
        \sum_{i=1}^{N}
        \bigl(y_i - f(\mathbf{x}_i;\boldsymbol{\theta})\bigr)^2
        -\frac{N}{2}\log(2\pi\sigma^2).
        \label{eq:log_likelihood_reg}
    \end{align}

The \textbf{maximum likelihood estimation} (MLE) searches for the values of $\boldsymbol{\theta}$ that maximize the likelihood function, or equivalently, the log-likelihood function~\cite[Sec.~2.5]{pawitan_all_2001}.
In the case of point-wise predictions and disregarding uncertainties, the optimization can be simplified to a least squares problem
    \begin{align*}
        \arg\max_\theta \ell_N(\boldsymbol{\theta})
        = \arg\min_\theta \frac{1}{2\sigma^2} \|\mathbf{y} - f(\mathbf{x};\boldsymbol{\theta})\|^2,
    \end{align*}
where $f(\mathbf{x} ;\boldsymbol{\theta}) = \bigl( f(\mathbf{x}_1; \boldsymbol{\theta}), \dots, f(\mathbf{x}_N;\boldsymbol{\theta}) \bigr)^\top$ denotes the vector of model predictions. 
To perform UQ, $\sigma^2$ must be estimated either by using an unbiased estimator \emph{a priori}, by optimizing it together with $\boldsymbol{\theta}$, or by estimating it with the MSE of that value of $\boldsymbol{\theta}$.
In the case of \textbf{heteroskedasticity}, the Gaussian noise distribution has a different variance $\sigma_i^2$ depending on each data point $\mathbf{x}_i$, with the corresponding log-likelihood~\cite{wu_variable_2012}
\begin{align*}
    \ell_N(\boldsymbol{\theta}) =
    -\frac{1}{2}
    \sum_{i=1}^{N}
    \frac{\bigl(y_i - f(\mathbf{x}_i;\boldsymbol{\theta})\bigr)^2}{\sigma_i^2}
    -\frac{1}{2}\sum_{i=1}^{N}
    \bigl(\log(2\pi\sigma^2_i)\bigr).
\end{align*}
    
If the variance for each data point is known through sensor specifications or repeated measurements, the optimization problem can be formulated as a weighted least squares problem. 
An alternative is to use a proxy function $g(\mathbf{x}_i; \gamma) \approx \sigma_i^2$ and optimize both $f$ and $g$ simultaneously, or to interchange between optimizing $\boldsymbol{\theta}$ and estimating $\sigma_i^2$ until convergence. 

The MLE is typically achieved through gradient-based optimization by minimizing the negative of the log-likelihood function as the objective function~\cite[Chap.~8]{murphy_probabilistic_2022}.
The gradient of the Gaussian log-likelihood is calculated as
\begin{align}
    \nabla_{\boldsymbol{\theta}} \ell_N(\boldsymbol{\theta}) &=
    \frac{1}{\sigma^2}
    J(\boldsymbol{\theta})^\top
    \bigl(\mathbf{y} - f(\mathbf{x};\boldsymbol{\theta}) \bigr),
\end{align}
where $J(\boldsymbol{\theta})$ is the Jacobian matrix of partial derivatives of the model outputs with respect to the parameters (for more details see the supplementary material, Appendix~\ref{sec:appendix_likelihood}).

The curvature of the parameter space can estimate the certainty associated with each parameter value and is calculated with the second-order derivatives in the Hessian matrix:
    \begin{align}
        &  H(\boldsymbol{\hat{\theta}}) = \nabla^2_{\boldsymbol{\theta}} \ell_N(\boldsymbol{\hat{\theta}}) \nonumber \\
         &= -\frac{1}{\sigma^2}
        \left(
        J(\boldsymbol{\hat{\theta}})^\top J(\boldsymbol{\hat{\theta}}) -
        \sum_{i=1}^{N} \bigl(y_i - f(\mathbf{x}_i;\boldsymbol{\hat{\theta}}) \bigr) \nabla^2_{\boldsymbol{\theta}} f(\mathbf{x}_i;\boldsymbol{\hat{\theta}}) \right) 
         \label{eq:hessian}
    \end{align}
    
Assuming that the parameter fitting algorithm has converged, the estimated parameters are close to an optimum. 
Under the additional assumption of i.i.d. Gaussian residuals $\mathcal{N}(0, \sigma^2)$, the last term in Eq.~\ref{eq:hessian} vanishes, resulting in the Gauss-Newton approximation of the Hessian~\cite[Sec.~10.2]{nocedal_numerical_1999}:
    \begin{align}
        H(\hat{\boldsymbol{\theta}})
        \approx
        -\frac{1}{\sigma^2}
        J(\hat{\boldsymbol{\theta}})^\top
        J(\hat{\boldsymbol{\theta}}).
    \end{align}

\begin{figure}[t!]
\centering
\begin{tikzpicture}
\begin{axis}[
    width=0.95\linewidth,
    height=0.43\linewidth,
    axis lines = left,
    xlabel = {$\theta$},
    ylabel = {$\ell(\theta)$},
    xtick = \empty,
    ytick = \empty,
    xmin = 0, xmax = 10,
    ymin = 0, ymax = 12,
    domain = 1:10,
    samples = 200,
    smooth,
    clip = false
]
\addplot [
    blue, 
    thick
] {10 * exp(-(x-2.5)^2 / 0.15) + 6 * exp(-(x-7)^6 / 5)};
\node[pin=90:{Global optimum}] at (axis cs:2.5, 10) {};
\node[pin=90:{Local optimum (plateau)}] at (axis cs:7, 6) {};
\end{axis}
\end{tikzpicture}
\vspace{-0.3cm}
\caption{Illustration of a multimodal likelihood search space with two optima, one being a plateau. Notice that the search space can have more than two optima, and even the global optimum can be a plateau.}\label{fig:likelihood}
\end{figure}
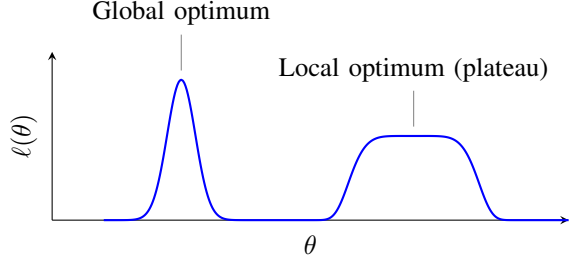

In regression analysis, the negative of the Hessian of the log-likelihood is also known as the observed Fisher information $I(\boldsymbol{\hat\theta)} \in \mathbb{R}^{d \times d}$, or simply observed information\footnote{See supplementary Appendix~\ref{sec:appendix_likelihood} for the \textit{expected} Fisher information.}. 
The diagonal elements of $H$ indicate how strongly the gradient of the log-likelihood changes in the direction of the parameter $\hat\theta_i$ (curvature), and the non-diagonal elements of the matrix indicate how the slopes with respect to two parameters $\hat\theta_i$ and $\hat\theta_j$ co-vary.
Intuitively, small Fisher information corresponds to a flat likelihood surface or plateau, such as the right peak in Fig.~\ref{fig:likelihood}.
This indicates an increased uncertainty about the true value of the parameters, as there exist many similar alternatives.
Plateaus can result from limited data or can be induced by the model structure, where only the former can be alleviated with more data. 
On the other hand, a sharp peak (left peak in Fig.~\ref{fig:likelihood}) indicates a precise parameter estimation around the optimum~\cite[Sec.~4.7.2]{murphy_probabilistic_2022}.

For nonlinear models, another challenge is the multimodality of the likelihood surface, which may contain many local and global optima potentially leading the optimization method into a suboptimal solution (as illustrated in Fig.~\ref{fig:likelihood}).
This can only be alleviated with more advanced optimization methods that, in turn, are more computationally expensive.
Importantly, quantifying parameter uncertainty does not resolve multimodality.
It only assesses the uncertainty around the particular optimum to which the parameter fitting algorithm has converged, and does not address the confidence in selecting the best among multiple optima. 

The next sections will cover the frequentist and Bayesian approaches, which depart from the aforementioned concepts.

\subsection{Frequentist approaches}
\label{sec:frequentist}

In frequentist statistics, a probability is interpreted as the long-run relative frequency of a measurable event in repeated identical experiments.
In this context, we assume that a true regression function exists together with its associated true parameter values, while the data comes from a random distribution.
The main goal of regression analysis in this context is to find a model that performs well over many hypothetical samples by using the likelihood function, i.e., the probability of observing the current data given a certain set of parameters~\cite[Sec.~4.7]{murphy_probabilistic_2022}.
Epistemic uncertainties in frequentist statistics are quantified by means of confidence intervals of the parameters and prediction intervals, as described in the following subsections.

\subsubsection{Basic concepts}

The measurements of uncertainty revolves around variance.
Three types of variances are relevant for regression models: 
the true (unknown) \textbf{variance} $\sigma^2$, 
the \textbf{estimated prediction variance} of the model $s^2$,
and the \textbf{variance of parameters} $S^2_{\boldsymbol{\theta}}$.
The estimated variance (as opposed to the true variance $\sigma^2$) and \textbf{residual standard error} $s$ of a regression model measure how much the data points deviate from the predictions.
This is calculated as
\begin{align*}
    s^2 = \frac{SSE}{N-d}, \qquad s = \sqrt{s^2}
\end{align*}
where $SSE$ is the sum of square errors. 
A large $s$ means that the data points are more spread around the mean, so the uncertainty of a point-wise prediction is high~\cite[Chap.~2]{bates_nonlinear_1988}. 

The same measurements can be calculated for the parameters $\boldsymbol{\theta}$ by using the information about the curvature given by $H$, resulting in the following equations for the variance matrix $S^2_{\boldsymbol{\theta}}$ and standard error $\operatorname{s}_{\theta_i}$ of the parameters:
\begin{align*}
    S^2_{\theta} (\boldsymbol{\hat{\theta}}) = s^2H(\boldsymbol{\hat{\theta}})^{-1}, \qquad
    \operatorname{s}_{\theta_i} = s \sqrt{H(\boldsymbol{\hat{\theta}})^{-1}_{i,i}}.
\end{align*}
The standard errors represent the units of uncertainties of the estimated parameters. 
For Gaussian distributions, they correspond to a $68.2\%$ confidence level (one standard deviation).
Small values indicate that we expect to observe little variation under repeated measurements, and thus a low parameter uncertainty~\cite[Chap.~2]{bates_nonlinear_1988}.

Another important concept in regression analysis is the \textbf{correlation matrix}, a standardized version of the covariance matrix with values in the range $[-1; 1]$, defined as
\begin{align*}
    C_{i,j} = \frac{H(\boldsymbol{\hat{\theta}})^{-1}_{i,j}}{\operatorname{s}_{\theta_i} \operatorname{s}_{\theta_j}}.
\end{align*}
High correlations between parameters indicate strong dependencies and potential redundancy, meaning that the model could be overparametrized~\cite[Chap.~2]{bates_nonlinear_1988}.

\subsubsection{Confidence and prediction intervals}
\label{sec:ci-pi}
In the context of regression, hypothesis tests can assess whether a regression parameter differs from zero (for details, see supplementary Appendix~\ref{sec:appendix_hypothesis}).
The \textbf{confidence interval of the parameters}~(CI) provide a similar perspective by giving a range of plausible parameter values instead of a binary answer whether the hypothesis can be rejected.
Over many repeated experiments, for $\alpha = 0.05$, $95\%$ of the constructed intervals using different samples drawn from the same population will contain the true parameter~\cite[Sec.~2.3]{bates_nonlinear_1988}.

\begin{figure}[t!]
\centering
\begin{tikzpicture}
\begin{axis}[
  no markers, 
  domain=-4:4, 
  samples=200, 
  axis lines*=left, 
  xlabel={$x$}, 
  every axis x label/.style={at=(current axis.right of origin), anchor=west},
  height=0.516\linewidth, 
  width=1.12\linewidth,
  xtick={-3,-2,-1,0,1,2,3},
  xticklabels={$\mu - 3\sigma$, $\mu- 2\sigma$, $\mu - \sigma$, $\mu$, $\mu+ \sigma$, $\mu+ 2\sigma$, $\mu + 3 \sigma$},
  ytick=\empty, 
  enlargelimits=false, 
  clip=false, 
  axis on top,
  hide y axis,
  declare function={gauss(\x,\mu,\sig)=1/(\sig*sqrt(2*pi))*exp(-((\x-\mu)^2)/(2*\sig^2));},
]

\addplot [fill=blue!10, draw=none, domain=-3:3] {gauss(x,0,1)} \closedcycle;
\addplot [fill=blue!20, draw=none, domain=-2:2] {gauss(x,0,1)} \closedcycle;
\addplot [fill=blue!40, draw=none, domain=-1:1] {gauss(x,0,1)} \closedcycle;

\addplot [very thick, blue!50!black] {gauss(x,0,1)};

\draw [thick, dashed, black!70] (axis cs:0,0) -- (axis cs:0,{gauss(0,0,1)})
    node[anchor=south] {$\mu$};

\node at (axis cs: 0.5, 0.15)  {34.1\%};
\node at (axis cs: -0.5, 0.15) {34.1\%};
\node at (axis cs: 1.5, 0.04)  {13.6\%};
\node at (axis cs: -1.5, 0.04) {13.6\%};

\end{axis}
\end{tikzpicture}
\vspace{-0.3cm}
\caption{Gaussian distribution mean, standard error, and significance levels (area under the curve).}\label{fig:gaussian}
\end{figure}
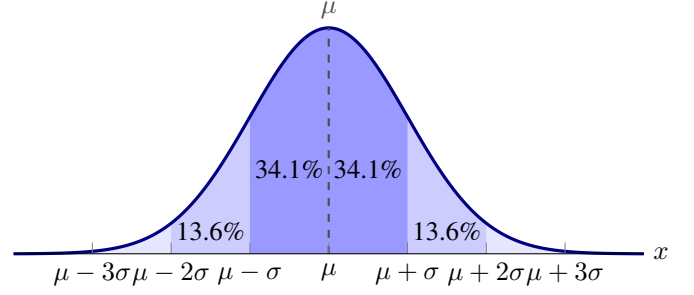

Given the point estimate $\hat\theta_i$ and the one unit of uncertainty $s_{\theta_i}$ for $\alpha = 0.682$ assuming Gaussian distribution (see Fig.~\ref{fig:gaussian}), this interval is defined by a lower and an upper bound
\begin{align*}
    \hat\theta_i \pm t_{\alpha/2,N-d} s_{\theta_i},
\end{align*}
where $t$ is a multiplicative factor from the Student-t distribution to achieve the desired significance level $\alpha$. 
Making the connection to hypothesis testing, if the value $0$ lies within the confidence interval around $\theta_i$, it means that there is not sufficient evidence that the parameter is different from 0 at the corresponding $\alpha$ and 
the null hypothesis cannot be rejected~\cite[Sec.~2.3]{bates_nonlinear_1988}. 

The \textbf{confidence interval of the predictions} is calculated analogously using the \textbf{delta method}~\cite{doob_limiting_1935} requiring the derivative vector\footnote{Note: this is the derivative of the model calculated at a single point $x^j$, not to be confused with the Jacobian $J$ of the likelihood function.} $v_j$ at any point $\mathbf{x}^j \in \mathcal{D}$~\cite[Chap.~2]{seber_nonlinear_1989}:
\begin{align*}
    f(\mathbf{x}^j; \boldsymbol{\hat{\theta}}) \pm s \sqrt{{v_j}^T H^{-1} v_j} \; t_{(\alpha/2,N-d)} \text{, with } v_j = \frac{\partial f(\mathbf{x}^j; \boldsymbol{\hat{\theta}})}{\partial \boldsymbol{\hat{\theta}}}
\end{align*}

For an arbitrary data point $x^j \notin \mathcal{D}$, the \textbf{prediction interval} accounts for an additional standard error as
\begin{align*}
    f(x^j, \boldsymbol{\hat{\theta}}) \pm s \sqrt{1 + {v_j}^T H^{-1} v_j} \; t_{\alpha/2,N-d}.
\end{align*}

A confidence interval constructs a range so that the true mean is included in $(1-\alpha) 100\%$ of repeated samples, while a prediction interval is constructed to contain the range in which future observations are expected to fall with the same long-run frequency.
Both can become narrower with larger sample size (reducing uncertainty about the mean), but the prediction interval will always remain wider than the confidence interval as it includes irreducible noise~\cite[Chap.~2]{seber_nonlinear_1989}. 

\subsubsection{Conformal prediction}\label{sec:conformalprediction}
\textbf{Conformal prediction}~(CP) is a framework used to construct prediction intervals with a guaranteed coverage. 
Unlike traditional methods that rely on distributional assumptions like normality, conformal prediction ensures that for a chosen confidence level $1-\alpha$, the resulting interval will contain the true future observation at least $(1-\alpha)100\%$ of the time, provided the data points are exchangeable.
Exchangeability in this context means that the joint distribution does not change when the order of the data is permuted, which is weaker than the i.i.d. constraint. 
It can be violated when the data inherently contains some dependence, such as temporal dependence for time-series data, or hierarchical dependence in grouped datasets. 
In the case of regression analysis, since i.i.d. is typically assumed, exchangeability is already included. 
Thus, instead of establishing a confidence level $\alpha$, conformal prediction defines a coverage. 
So, if we ask for a $95\%$ coverage, it means that the probability that the set contains the true value is almost exactly $95\%$, given that it involves empirical evaluation of unseen data~\cite{angelopoulos_conformal_2023}. 

The procedure to compute a prediction interval using conformal prediction works as follows: 
First, the data is split into two dataset, namely training set $\mathcal{D}_{train}$ and calibration set $\mathcal{D}_{cal}$. 
The former is used to train the model, and the latter to constructed the prediction interval. 
After the model is trained, the calibration residuals or non-conformity scores are computed as 
\begin{align*}
    r_i = \lvert y_i - \hat{f}(x_i) \rvert, \quad i \in \mathcal{D}_{\text{cal}}.
\end{align*}

In regression settings, the absolute error is the most commonly used nonconformity score.
This score is employed to estimate the empirical quartile $q_{1-\alpha}$, which is selected as the $\left\lceil (1-\alpha)(m+1) \right\rceil$-th smallest residual $r_i,~i\in \mathcal{D}_{\text{cal}}$, where $m = \lvert\mathcal{D}_{\text{cal}}\rvert $.
The resulting prediction interval for new observations $x_*$ can then be computed as
\begin{align*}
    \Gamma(x_*) =
\big[\, \hat{f}(x_*) - q_{1-\alpha},\;
       \hat{f}(x_*) + q_{1-\alpha} \,\big].
\end{align*}

For example, if a dataset has 247 samples and for $\alpha = 0.05$, the residual value at rank $\left\lceil 235.6 \right\rceil = 236$ would determine the quartile of the prediction interval. 
Because of the exchangeability property, the rank of of the score of any new data point will have a uniform probability of falling in between any two ranks in the calibration set. 
So the probability that the new rank is less than or equal to $q_{1-\alpha}$ is $1 - \alpha$~\cite[Appendix D]{angelopoulos_conformal_2023}.

To summarize, CP has three key properties:
\begin{itemize}
    \item Finite-sample coverage: guarantees the coverage even with small datasets
    \item Model-agnostic: this procedure does not depend on the inner working of the model
    \item Distribution-free: no prior distribution is assumed
\end{itemize}

While prediction intervals and conformal prediction answer different questions, there are also practical differences that must be taken into consideration before choosing one of the techniques.
Prediction intervals work with any amount of data since they rely on the Gaussian distribution assumption.
Also, the intervals have a statistical interpretation since they are based on the Fisher information, which takes the parameters values into consideration.
On the other hand, if the assumption about a distribution is wrong, the $(1-\alpha)100\%$ may be lower than requested, which can be alleviated with other techniques such as likelihood profiles.
As the intervals are more narrow, they can be too optimistic and lead to misguided decisions.

Regarding conformal prediction, the property of guaranteed coverage and being distribution free are the main advantages.
However, requires enough data to perform the split into training and calibration sets.
While the coverage is guaranteed, the interval can become very wide if the number of samples is low, which decreases its practical applicability.

\subsection{Bayesian methods}
\label{sec:bayesian}

In frequentist statistics, the parameters are treated as fixed but unknown, and probability describes the variability of the data given those parameters. 
In Bayesian statistics, to the contrary, the parameters are a random variable following a certain distribution, and inference is based on the distribution of the parameters given the observed data~\cite[Chap.~4.6]{murphy_probabilistic_2022}.

\subsubsection{Bayesian inference}
\label{sec:bayes_inference}
In Bayesian regression, the \textbf{posterior distribution} $p(h \mid \mathcal{D})$ describes the probability of a hypothesis $h$ given the data $\mathcal{D}$, following the Bayes equation
\begin{align}
    p(h \mid \mathcal{D}) = \frac{p(h)\, p(\mathcal{D} \mid h)}{p(\mathcal{D})}
    \propto p(h)\, p(\mathcal{D} \mid h),
    \label{eq:bayes}
\end{align}
where $p(\mathcal{D} \mid h)$ is the likelihood of $h$, $p(h)$ is the \textbf{prior probability} of the hypothesis, and $p(\mathcal{D})$ is the \textbf{marginal likelihood} which acts as a normalizing factor ensuring that the posterior will add up to one\footnote{The marginal likelihood (or evidence) is obtained by integrating out the parameter values: $p(D) = \int_{\boldsymbol{\theta}} p(D,\boldsymbol{\theta)} \, d\boldsymbol{\theta} = \int_{\boldsymbol{\theta}} p(D | \boldsymbol{\theta}) p(\boldsymbol{\theta}) \, d\boldsymbol{\theta}$.}~\cite[Sec.1.3]{gelman_bayesian_2013}.
In this section, we focus on a traditional regression analysis case, where the model structure is fixed, and $p(h)$ refers to the prior over the parameters.
In later sections, some surveyed papers treat $p(h)$ as the probability distribution over symbolic models, but the underlying techniques introduced in this section still apply.

To calculate the posterior, the prior belief is updated with the evidence from the observed data via the likelihood function $p(\mathcal{D}\mid h)$.
Due to the intractability of the marginal likelihood, this process is often performed using \textbf{Markov Chain Monte Carlo}~(MCMC) sampling~\cite[Chap.~11]{gelman_bayesian_2013}. 
In MCMC, the posterior distribution is only required up to proportionality, i.e., the last part in Eq.~\ref{eq:bayes}, disregarding the marginal likelihood.
Thus, the algorithm evaluates the posterior density up to a normalizing constant.

A Markov chain is a sequence of random variables $\left\{x^{(1)}, x^{(2)}, \ldots, x^{(k)}\right\}$ such that the $k$-th value in the sequence only depends on its predecessor\footnote{We adopt the notion of parenthesized superscript $\Box^{(t)}$ as the state $t$ in the Markov chain.}:
\begin{align*}
    P(X^{(t)} &= x^{(t)} \mid X^{(t-1)} = x^{(t-1)},\ldots,X^{(1)} = x^{(1)}) \\
    &= P(X^{(t)} = x^{(t)} \mid X^{(t-1)} = x^{(t-1)}).
\end{align*}

Monte Carlo sampling estimates statistical properties by repeated sampling when an analytical solutions is unavailable or expensive to calculate. 
These two concepts can be combined as a technique that performs repeated sampling (Monte Carlo), where the next sample is only dependent on the previous sample (Markov chain). 
This creates a chain of random variables that can be used to estimate unknown probability distributions or to update prior beliefs.
The general MCMC algorithm is \textbf{Metropolis-Hastings}~(MH), composed of two steps: proposal sampling, and acceptance or rejection~\cite[Sec.~11.2]{gelman_bayesian_2013}.

Using a regression example, the MH algorithm starts with an initial set of model parameters $\boldsymbol{\theta}$ and variance of the data $\sigma^2$, typically drawn from their prior distributions.
In each iteration, new parameter values are proposed based on the previous ones, for example $\theta_i^{(t)} \sim \mathcal{N}(\theta_i^{(t-1)}, \sigma_{\theta}^2)$ assuming a Gaussian proposal distribution. 
Likewise, a new value is proposed, typically on the log-scale, for $\log \sigma^2 \sim \mathcal{N}(\log {\sigma^2}^{(t-1)}, s^2)$, which ensures that $\sigma^2>0$ while allowing samples from an unconstrained space, and $s^2$ is a hyperparameter that controls the step size of the proposal distribution~\cite{browne_mcmc_2006}. 
Finally, in our example the proposal is accepted with the probability 

\begin{align*}
r = &\min\left\{1,\;
\frac{
p(\mathcal{D} \mid \boldsymbol{\theta}^{(t)}, {\sigma^2}^{(t)}) \, p({\sigma^2}^{(t)})
}{
p(\mathcal{D} \mid \boldsymbol{\theta}^{(t-1)}, {\sigma^2}^{(t-1)}) \, p({\sigma^2}^{(t-1)})
}
\cdot q_r \right\},\\
&\text{where }
q_r = \frac{
q(\boldsymbol{\theta}^{(t-1)}, {\sigma^2}^{(t-1)} \mid \boldsymbol{\theta}^{(t)}, {\sigma^2}^{(t)})
}{
q(\boldsymbol{\theta}^{(t)}, {\sigma^2}^{(t)} \mid \boldsymbol{\theta}^{(t-1)}, {\sigma^2}^{(t-1)})
},
\end{align*}
assuming a uniform prior for $\boldsymbol{\theta}$, and the probability $q(t\mid t')$ of reaching a certain state $t$ from the state $t'$. 
Otherwise, the previous state is kept.
For symmetric proposal distributions such as the Gaussian proposal, the term $q_r$ cancels out. 
Importantly, the proposal and prior distributions are not the same in MCMC: the prior defines the target posterior distribution, and the proposal determines how states are sampled.

One disadvantage of the MH algorithm is that simple proposal schemes may update multiple parameters independently from each other, which can lead to inefficient sampling and slow convergence in high-dimensional or highly correlated parameter spaces.
Gibbs sampling addresses this problem by sampling each $\theta_i^{(t)}$ from the full conditional distribution $p(\theta_i \mid \boldsymbol{\theta_{j \neq i}^{(t-1)}}, \mathcal{D})$.
This always returns an acceptance rate of $1$, since samples are drawn from the target conditional distributions~\cite[Sec.~11.1]{gelman_bayesian_2013}.
In nonlinear regression and specifically SR, the conditional distributions are typically not available in closed form, and thus most of the works use simple proposal schemes.
When only some of the conditional probabilities are available, hybrid methods such as Metropolis-within-Gibbs can be used to apply the most optimal step accordingly~\cite{gilks_adaptive_1995}.

Common choices of parameter priors are the Gaussian distribution, acting like  $l_2$ regularization, Laplace distribution, acting like $l_1$, and uniform distribution. 
For the noise variance $\sigma^2$, a common distribution is inverse gamma $\text{IG}(\alpha, \beta)$, which is restricted to the range $(0, \infty)$ and conjugate with the Gaussian likelihood.
A uniform prior over the range $(-\infty, \infty)$ is called an \textbf{improper prior}, as it does not integrate to $1$.

\subsubsection{Credible Interval}
\label{sec:credible_int}
\begin{figure}[t]
\centering
\begin{tikzpicture}[
    declare function={
        skewed(\x) = ((\x > 0) * (\x^2 * exp(-\x))) / 0.541;
    }]
\begin{axis}[
    width=\linewidth, 
    height=0.55\linewidth,
    axis lines=left,
    xlabel={$\theta$},
    ylabel={Probability density},
    xlabel style= {at={(0.5,0)}, anchor=north},
    ylabel style= {at={(0,0.5)}, anchor=south},
    domain=0:10,
    samples=200,
    ticks=none,
    enlargelimits=false,
    clip=false,
    ymin=0, ymax=1.2,
    xmin=0, xmax=10.5
]

\def\hpdlevel{0.35}

\addplot [name path=curve, very thick, blue!50!black] {skewed(x)};

\path [name path=bottom] (axis cs:0,0) -- (axis cs:10,0);

\addplot [fill=blue!10, opacity=0.8] fill between [of=curve and bottom];

\fill [blue!30, opacity=0.5] (axis cs:0.95, 0) rectangle (axis cs:4.6, 1.2);
\addplot [very thick, blue!50!black] {skewed(x)};

\draw [very thick] (axis cs:0.95, 0) -- (axis cs:0.95, 1.2);
\draw [very thick] (axis cs:4.6, 0) -- (axis cs:4.6, 1.2);
\draw [thick, dashed] (axis cs:0.95, 1.05) -- (axis cs:4.6, 1.05);

\node at (axis cs: 2.7, 0.5) {\small{90\%}};
\node [anchor=south east] at (axis cs: 1.12, 0.01) {\small{2\%}};
\node [anchor=south west] at (axis cs: 4.8, 0.01) {\small{8\%}};
\node [anchor=south east] at (axis cs: 0.92, 1) {\normalsize $l$};
\node [anchor=south west] at (axis cs: 4.55, 1) {\normalsize $u$};
\end{axis}
\end{tikzpicture}
\vspace{-0.5cm}
\caption{Credible interval for $1 - \alpha = 0.9$. The lower and upper bounds is defined by the darker shaded region containing the $90\%$ of the mass.}\label{fig:credible}
\end{figure}
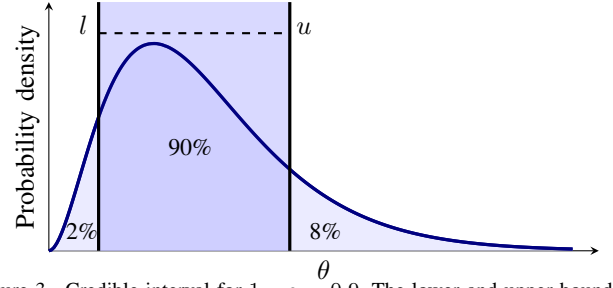

Regarding UQ, the \textbf{credible interval} is the Bayesian analogue to the frequentist confidence interval.
It is a region of the parameter space that contains $(1-\alpha)100\%$ of the posterior probability mass (see Fig.~\ref{fig:credible})~\cite[Sec.~2.3]{gelman_bayesian_2013}.
The goal is to identify lower and upper bounds $[l, u]$ so that
\begin{align}  
    \int_l^u p(\theta \mid \mathcal{D}) \, d\theta = 1 - \alpha.
    \label{eq:credible_int}
\end{align}

When the posterior can be analytically derived, the credible interval contains the true parameter with a probability of $(1-\alpha)100\%$, given the Bayesian model and the observed data. 
Thus, credible intervals allow for a direct probabilistic interpretation.
In contrast, a confidence interval only guarantees that $(1-\alpha)100\%$ of the intervals constructed from repeated sampling from the same distribution contains the true parameter. 
For a single dataset, whether the true mean is inside or outside the CI cannot be verified in practice. 

The calculation of the credible interval in Eq.~\ref{eq:credible_int} differs between textbook-like Bayesian methods and practical implementation in regression algorithms.
In textbook settings, the selection of conjugate priors enables the closed-form calculation of the posterior, where conjugacy means that the posterior belongs to the same distributional family as the prior when combined with the likelihood.
This known posterior family allows for a closed-form evaluation of integrals over the parameter space. 
In SR, such inferences are however rarely feasible and the interval over the posterior is thus not tractable.  
Instead, it can be empirically estimated as an average over the samples drawn by the MCMC algorithm. 
Typically, the narrowest credible interval, i.e., the region with the highest posterior density, is preferred. 
It can be approximated by sorting the posterior samples and using a sliding window to select the shortest interval containing $(1-\alpha)100\%$ of samples.

Next to the credible interval for parameters, we can also calculate the \textbf{credible interval for the posterior mean} of any point $x^{j}$ in the training data with the same procedure.
Using the chain of MCMC-sampled parameters $\{\boldsymbol{\theta}^{(t)}\}_{t=1 \ldots N}$, we generate the set $\{y^j = f(x^j, \boldsymbol{\theta}^{(t)})\}_{t=1 \ldots N}$, i.e., a set of predictions over different parameter values drawn from the posterior~$p(\boldsymbol{\theta} \mid \mathcal{D})$.

Moreover, the \textbf{posterior prediction interval} for a new observation $x_*$ is computed as 
\begin{align*}
    p(y_* \mid x_*, \mathcal{D}) = \int p(y_* \mid x_*, \boldsymbol{\theta}, \sigma^2)\, p(\boldsymbol{\theta}\mid \mathcal{D})\, d\boldsymbol{\theta} \,d\sigma^2, 
\end{align*}
where $p(y_* \mid x_*, \boldsymbol{\theta},\sigma^2)$ is the likelihood of an output $y_*$ given the new observation and a certain $\theta_i$ under the assumed noise distribution $\epsilon \sim \mathcal{N}(0, \sigma^2)$, and $p(\theta \mid \mathcal{D})$ is simply the posterior distribution.
This integral is again often intractable in practice, and thus approximated using the samples $\{(\boldsymbol{\theta}^{(t)}, \sigma^{2^{(t)}})\}_{t=1 \ldots N}$ obtained during the MCMC procedure, which generates the predictions
\begin{align*}
    \{y_*^{t} \sim\mathcal{N}(f(x_*, \boldsymbol{\theta}^{(t)}), \sigma^{2^{(t)}}\}_{t=1 \ldots N}.
\end{align*}
The posterior predictive interval is then obtained from these samples using the sliding-window procedure to find the region of highest posterior density.
The additional step of sampling from a Gaussian distribution in the previous equation is used to account for the aleatoric uncertainty of the data, resulting in wider prediction intervals compared to the credible intervals for the posterior mean.

\subsection{Model selection under uncertainties}
\label{sec:model_selection}
\textbf{Model selection} plays an important role in SR, as typically one among many alternative hypotheses needs to be selected~\cite[Sec.~5.2]{murphy_probabilistic_2022}. 
It is not a trivial task, as both aleatoric and epistemic uncertainties add layers of complexity to the selection process.
Aleatoric uncertainty can lead to overfitting, as an apparently perfect model at the MLE might fit the noise, and the observed error does not clearly separate model inadequacy from measurement noise.
Epistemic uncertainty is connected to the identifiability problem where many models fit the data equally well and cannot be distinguished without further information. 
This section introduces more advanced model selection principles under uncertainty commonly used in machine learning.
The basic concepts, namely cross validation, Akaike information criterion~(AIC) and Bayesian information criterion~(BIC), are detailed in the supplementary material, Appendix~\ref{sec:appendix_modelselection}.

\subsubsection{Bayes Factor and Fractional Bayes Factor}\label{sec:fbf}

\begin{figure}[t!]
\centering
\begin{tikzpicture}

\begin{groupplot}[
    group style={
        group size=2 by 1,
        horizontal sep=0.3cm
    },
    width=5.2cm,
    height=4.3cm,
    xlabel={$\theta$},
    xlabel style={yshift=0.05cm},
    ylabel={Density},
    ymin = 0,
    ymax = 0.7,
    domain=-4:4,
    samples=200,
]

\nextgroupplot[
    title={Simple Model},
    legend style={at={(1.0,-0.3)}, anchor=north, legend columns=3, font=\scriptsize, draw=none, fill=none},
]

\addplot[thick, black]
    {1/sqrt(2*pi*0.7^2) * exp(-(x-1)^2/(2*0.7^2))};
\addlegendentry{Likelihood}

\addplot[thick, blue, dashed]
    {1/sqrt(2*pi*0.8^2) * exp(-(x-0)^2/(2*0.8^2))};
\addlegendentry{Prior}

\addplot[thick, red]
    {1/sqrt(2*pi*0.7^2) * exp(-(x-1)^2/(2*0.7^2)) * 1/sqrt(2*pi*0.8^2) * exp(-(x-0)^2/(2*0.8^2))};
\addlegendentry{Posterior}

\nextgroupplot[
    title={Complex Model},
    ylabel={},          
    yticklabels={}   
]

\addplot[thick, black]
    {1/sqrt(2*pi*0.7^2) * exp(-(x-1)^2/(2*0.7^2))};

\addplot[thick, blue, dashed]
    {1/sqrt(2*pi*2.5^2) * exp(-(x-0)^2/(2*2.5^2))};

\addplot[thick, red]
    {1/sqrt(2*pi*0.7^2) * exp(-(x-1)^2/(2*0.7^2)) * 1/sqrt(2*pi*2.5^2) * exp(-(x-0)^2/(2*2.5^2))};

\end{groupplot}

\end{tikzpicture}
\vspace{-0.5cm}
\caption{Comparison of prior (dashed blue line), likelihood (solid black line), and posterior (solid red line) under simple and complex model assumptions. This illustrates the influence of the model complexity (number of parameters) on the value of $q_i$, which is the integral over the posterior.}
\label{fig:bayes_compare}
\end{figure}
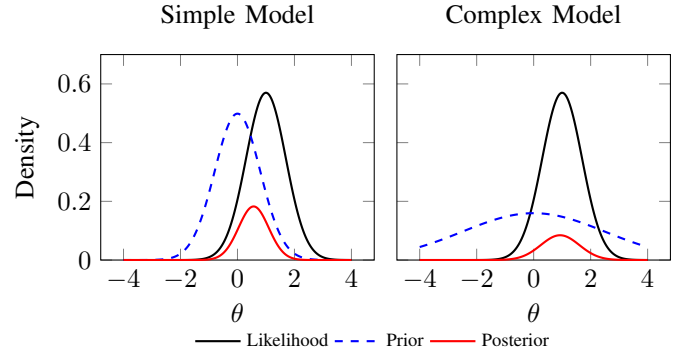

The \textbf{Bayes factor}~(BF) compares two hypothesis using their posterior odds as
\begin{align*}
    \frac{p(\boldsymbol{\theta}_1 \mid \mathcal{D}, h_1)}{p(\boldsymbol{\theta}_2 \mid \mathcal{D}, h_2)} &= \frac{p(\boldsymbol{\theta}_1 \mid h_1) q_1(\mathcal{D})}{p(\boldsymbol{\theta}_2 \mid h_2) q_2(\mathcal{D})} 
    = \frac{p(\boldsymbol{\theta}_1 \mid h_1)}{p(\boldsymbol{\theta}_2 \mid h_2)} B(\mathcal{D}),
\end{align*}
where $B(\mathcal{D})$ is the BF in favor of $h_1$ against $h_2$, and 
\begin{align*}
    q_i(\mathcal{D}) = \int{p(\boldsymbol{\theta}_i \mid h_i) p(\mathcal{D} \mid \boldsymbol{\theta}_i, h_i) d \boldsymbol{\theta}_i}
\end{align*}
is the marginal density of the data over hypothesis $h_i$~\cite{ohagan_fractional_1995}.
In the common case where all hypothesis are equally likely, the posterior odds can be reduced to the Bayes factor $B(\mathcal{D})$.
When the likelihood functions are similar for both hypothesis, the difference of values in $q_i$ will be determined by their parameter priors.
Models with more parameters spread the probability mass over a larger parameter space, which reduces the marginal likelihood.
As a result, the simpler model will have a larger $q_i$ value and is preferred (see Fig.~\ref{fig:bayes_compare}).

One caveat with the Bayes factor is that Bayesian regression commonly uses an improper prior, such as a uniform distribution $p(\boldsymbol{\theta}_i \mid h_i) \propto 1$, which makes its integral over the parameter space infinity and the marginal likelihood undefined.
Alternatively, the \textbf{fractional Bayes factor}~(FBF) corrects the improper prior using a fraction of the likelihood.
The fractional prior $p_\gamma$ is calculated as
\begin{align}
    p_\gamma(\boldsymbol{\theta}_i \mid h_i) = \frac{p(\boldsymbol{\theta}_i \mid h_i) p(\mathcal{D} \mid \boldsymbol{\theta}_i, h_i)^\gamma}{\int{p(\boldsymbol{\theta}_i \mid h_i) p(\mathcal{D} \mid \boldsymbol{\theta}_i, h_i)^\gamma} d \boldsymbol{\theta}_i},
    \label{eq:fractional-prior}
\end{align}
where $0 < \gamma < 1$ defines the portion of the likelihood used to generate a proper prior, while the remaining $1 - \gamma$ are used to calculate the FBF~\cite{ohagan_fractional_1995}.
To illustrate the correction of an improper prior for a linear model, consider the Gaussian distribution of the likelihood, which quickly approaches $0$ as $\lvert {\theta}_i\rvert \rightarrow \infty$. 
When the prior is multiplied with a fraction of the likelihood as in Eq.~\ref{eq:fractional-prior}, it adjusts the probability of the less likely values of $\boldsymbol{\theta}$ to $0$, making the prior proper.
The value of $\gamma$ should be chosen as small as possible to leave sufficient information for computing the marginal likelihood.
It is obtained by replacing the prior with the corrected prior and using the remaining fraction $(1-\gamma)$ of the likelihood~\cite{ohagan_fractional_1995}:
\begin{align}
    q_i(\mathcal{D,\gamma}) &= \int{p_\gamma(\boldsymbol{\theta}_i \mid h_i) p(\mathcal{D} \mid \boldsymbol{\theta}_i, h_i)^{(1-\gamma)} d \boldsymbol{\theta}_i} \nonumber \\
    &= \frac{\int{p(\boldsymbol{\theta}_i \mid h_i) p(\mathcal{D} \mid \boldsymbol{\theta}_i, h_i) d \boldsymbol{\theta}_i}}{\int{p(\boldsymbol{\theta}_i \mid h_i) p(\mathcal{D} \mid \boldsymbol{\theta}_i, h_i)^\gamma} d \boldsymbol{\theta}_i}.
    \label{eq:fbf}
\end{align}

Commonly, $\gamma = \frac{d}{N}$ is used for linear models~\cite[Chap.~6]{ohagan_fractional_1995}. 
The resulting integral can be approximated using the Laplace approximation using the MLE, i.e., the likelihood at $\hat{\boldsymbol{\theta}}$:
\begin{align*}
    \int{e^{N f(\mathbf{x};\boldsymbol{\theta})} d \theta} \approx e^{N f(\mathbf{x};\hat{\boldsymbol{\theta}})} 2 \pi^\frac{d}{2}|N H(\hat{\boldsymbol{\theta}})|^{-0.5}.
\end{align*}

Rewriting the numerator in Eq.~\ref{eq:fbf} in terms of the log-likelihood, i.e., $p(\mathcal{D} \mid \boldsymbol{\theta}_i, h_i) = \exp(\ell_N(\boldsymbol{\theta}))$, and applying Laplace approximation results in:
\begin{align*}
    &\int{p(\boldsymbol{\theta}_i \mid h_i) p(\mathcal{D} \mid \boldsymbol{\theta}_i, h_i) d \boldsymbol{\theta}_i}  
    = \int{p(\boldsymbol{\theta}_i \mid h_i) e^{\ell_N(\boldsymbol{\theta}_i)} d \boldsymbol{\theta}_i} \\
    &\approx p(\hat{\boldsymbol{\theta}}_i \mid h_i) p(\mathcal{D} \mid \hat{\boldsymbol{\theta}}_i, h_i) (2\pi)^{d/2} |N H(\hat{\boldsymbol{\theta}})|^{-0.5} \\ 
    &= p(\hat{\boldsymbol{\theta}}_i \mid h_i) p(\mathcal{D} \mid \hat{\boldsymbol{\theta}}_i, h_i) (2\pi)^{d/2} N^{-d/2}|H(\hat{\boldsymbol{\theta}})|^{-0.5} .
\end{align*}

Finally, the marginal density becomes
\begin{align}
    q_i(\mathcal{D,\gamma}) &= \frac{p(\hat{\boldsymbol{\theta}}_i \mid h_i) p(\mathcal{D} \mid \hat{\boldsymbol{\theta}}_i, h_i) (2\pi)^{\frac{d}{2}} N^{-\frac{d}{2}}|H(\hat{\boldsymbol{\theta}})|^{-0.5}}{p(\hat{\boldsymbol{\theta}}_i \mid h_i) p(\mathcal{D} \mid \hat{\boldsymbol{\theta}}_i, h_i)^\gamma (2\pi)^{\frac{d}{2}} (\gamma N)^{-\frac{d}{2}}|H(\hat{\boldsymbol{\theta}})|^{-0.5}} \nonumber \\
    &= p(\mathcal{D} \mid \hat{\boldsymbol{\theta}}_i, h_i)^{1 - \gamma} \gamma^{\frac{d}{2}}, \label{eq:fbf-final}
\end{align}
thus solely depending on the likelihood, the number of parameters $d$, and the fraction $\gamma$~\cite{ohagan_fractional_1995}.

\subsubsection{Minimum Description Length Principle}

The \textbf{minimum description length}~(MDL)
~\cite{hansen_model_2001, rissanen_modeling_1978} states that a model should accurately describe the data using as few symbols as possible.
According to MDL, a dataset can be represented most effectively by the function that achieves the greatest compression, minimizing the amount of information required to reconstruct or communicate the data through that functional form.
The codelength $L$ measures the units of information required to express the data, often as bits or nats.
It is composed of an error and a complexity:
\begin{align}
    L(\mathcal{D}) = L(h) + L(\mathcal{D} \mid h),
    \label{eq:MDL}
\end{align}
$L(\mathcal{D} \mid h)$ is the codelength of the data given a hypothesis, often expressed as the negative log-likelihood at the estimated optimum, so that $L(\mathcal{D} \mid h) = - \ell_N(\hat{\boldsymbol{\theta}})$.
$L(h)$ is the codelength of the hypothesis expressing its complexity, typically considering multiple aspects of complexity. 

\section{Survey of Uncertainty-Related Methods in SR}

\subsection{Methodology}

To ensure a comprehensive survey of UQ in SR, we searched the related papers using three different sources: DBLP\footnote{\texttt{\url{https://dblp.org/}}}, The Genetic Programming Bibliography\footnote{\texttt{\url{https://gpbib.cs.ucl.ac.uk/}}}, and Google Scholar\footnote{\texttt{\url{https://scholar.google.com/}}}.
As the focus is on SR, we searched for the following keywords together with the term "symbolic regression", always in quotes:
``uncertainties'',
``uncertainty'',
``uncertainty quantification'',
``fisher information'',
``hessian'',
``epistemic'',
``aleatoric'',
``conformal prediction'',
``bayesian'',
``bayes'',
``minimum description length'',
``MDL'',
``AIC'',
``BIC'',
``model selection'',
``regularization'',
``noisy data'',
``noise'',
``bias-variance'',
``statistical machine learning''.

From the returned articles, we considered those that either proposed or adapted the use of any UQ method, methods requiring UQ (e.g., using the MDL principle for model selection), or establish prerequisites that enable UQ (e.g., Bayesian posterior distribution), in the context of SR. 
To keep the content of this paper focused, articles employing well established statistical ML concepts in SR, such as cross validation or the straight forward use of AIC and BIC for model selection, were excluded. 
The topic of generalization in SR is also disregarded from this survey, as it primarily focuses on predictive performance on unseen data rather than explicit mechanisms for UQ.
Furthermore, research on SR in dynamic environments (sometimes called ``uncertain environments'') falls outside the scope of this work.
Moreover, we decided to discard articles with incomplete information about the main UQ-related approach, which could not be resolved when contacting the authors. 

It is also important to clarify the relationship between UQ, as studied in this article, and robustness, which is a more established concept in evolutionary computation and not explicitly addressed in this article. 
While robustness assesses the sensitivity of models or algorithms to perturbations in the inputs, such as outliers in the data or corrupted inputs, UQ aims to explicitly quantify the uncertainty in predictions, parameters, and model structures.
UQ provides uncertainty estimates rather than guaranteeing stable predictions and thus supports reliable decision making under incomplete information.
Both concepts are related, but distinct: Improving robustness can also affect UQ, as for example removing outliers from the data will improve the reliability of the model, but, if wrongly dropping genuine data points, can result in unreliable uncertainty estimates. 
Despite their relation, this article focuses specifically on UQ methods due to their under-representation in the SR literature.

Overall, we focused on papers from peer-reviewed journals and conferences, but also included recent preprints from arXiv when the proposed methods appeared promising and the peer-review process was likely not yet completed.
We identified 11 relevant UQ-related articles focusing on Bayesian SR, 4 targeting non-Bayesian (but not only frequentist) methods, and 3 for model selection, which will be presented subsequently.
\subsection{Bayesian methods}
\subsubsection{Bayesian SR with Sequential Monte Carlo} 
\label{sec:bomarito_leser_papers}
Four consecutive publications ~\cite{bomarito_bayesian_2022,bomarito_automated_2023,leser_comparing_2024,bomarito_bayesian_2026} proposed and iteratively improved a Bayesian framework for SR that accounts for noisy data and enables UQ of the discovered models.
Within the evolutionary process and compared to conventional SR methods, this approach introduces changes at two steps~\cite{bomarito_bayesian_2022,bomarito_automated_2023}:
First, a probabilistic reformulation of the local optimization of parameters, and second, during the selection phase, where a replacement probability is computed from the marginal likelihood.
Using Bayes' theorem, the posterior probability of the parameters given the hypothesis $h$ is formulated as
\begin{align*}
p(\boldsymbol{\theta} \mid \mathcal{D}, h) 
&= \frac{p(\mathcal{D} \mid \boldsymbol{\theta}, h)\,p(\boldsymbol{\theta} \mid h)}{p(\mathcal{D} \mid h)}, 
\end{align*}
where $p(\mathcal{D} \mid h) = \int_{\mathbb{R}^d} p(\mathcal{D} \mid \boldsymbol{\theta}, h)\,p(\boldsymbol{\theta} \mid h)\, d\boldsymbol{\theta}$ is marginalized over the parameter space.
While Bayesian methods rely on prior information, the presented Bayesian GPSR approach uses an improper uniform prior $p(\boldsymbol{\theta} \mid h) \propto 1 $, 
which requires the FBF $\frac{q_1(\mathcal{D},\gamma)}{q_2(\mathcal{D},\gamma)}$ for model comparison and selection (see Eq.~\ref{eq:fbf} and Sec.~\ref{sec:fbf} for the derivation). 

Accurately estimating the integrals over the parameters in $q_i(\mathcal{D},\gamma)$ is challenging due to the intractability of the normalization constant. 
\citet{bomarito_bayesian_2022} propose a \textbf{sequential Monte Carlo}~(SMC) approach to estimate the integrals without assuming a parametric form of the posterior.
This method has been refined over a sequence of papers, where we focus on the most mature implementation~\cite{leser_comparing_2024}, which compares SMC with the Laplace approximation as proposed in~\cite{bartlett_priors_2023}.
For numerical purposes, they propose to compute the normalized marginal log likelihood~(NMLL) as $\log(q_i(\gamma))$.

The Laplace approximation of the NMLL assumes that the posterior distribution is a unimodal multivariate normal distribution~(MVN), centered at the \textbf{maximum a posteriori} point~(MAP) $\boldsymbol{\theta}^* = \arg \max_{\boldsymbol{\theta}} p (\boldsymbol{\theta} |\mathcal{D}, h)$.
This distribution has only one peak, and, in the case of a uniform prior, the MAP can be estimated as the MLE $\boldsymbol{\theta}^* = \boldsymbol{\hat{\theta}}$. 
Applying a logarithm to Eq.~\ref{eq:fbf-final}, \citet{leser_comparing_2024} compute the Laplace approximation of the NMLL for a uniform prior as
\begin{align*}
    \log \hat{q}_j(\gamma) = (1 - \gamma)\,\log p(\mathcal{D} \mid \boldsymbol{\hat{\theta}}, h_j) + \frac{d}{2}\,\log \gamma.
\end{align*}
Since finding the MLE with a nonlinear optimization method is more efficient than MCMC sampling, the Laplace approximation simplifies the evaluation of the integrals in $q_i$. 
However, posterior distributions are not always unimodal, which limits the accuracy of the Laplace approximation that only captures the curvature around a single optimum.

As an alternative to MCMC, SMC is a sample-based approach that can approximate arbitrary distributions using a sufficient number of weighted particles rather than a single Markov chain.
The main idea is that the posterior distribution is often concentrated in relatively small regions of the parameter space and close to zero in the remaining space. 
Thus, SMC avoids wasting samples in low-probability areas by 
letting the particle population explore and concentrate in high-probability regions through reweighting and resampling.
The posterior is approximated by sequentially moving from a distribution that is easy to sample from (such as the prior distribution), towards the posterior by introducing a larger portion of the likelihood at every step:
\begin{align}
    p_t
    \propto 
p(\boldsymbol{\theta}\mid h)\,
p(\mathcal{D}\mid \boldsymbol{\theta},h)^{\Phi_t},
\label{eq:intermediate_p}
\end{align}
\text{ with }
$0=\Phi_0 < \Phi_1 < \dots < \Phi_T=1$.

Conveniently, SMC naturally provides both terms required for $q_i$, if $\Phi_{t^*}= \gamma$ at an intermediate time step $t^*$.
The marginal likelihood can then be estimated at step $t=t^*$ (corresponding to the denominator in Eq.~\ref{eq:fbf}) and $t=1$ (corresponding to the numerator).
The SMC algorithms are implemented in the \texttt{SMCPy}\footnote{\texttt{\url{https://github.com/nasa/smcpy}}} package with details described in the following.

At $t = 0$, a set of $N_p$ particles is sampled from a proposal distribution $\boldsymbol{\theta}_{i,0} \sim g(\boldsymbol{\theta} \mid h)$ for $i = 1, \dots, N_p$. 
Importantly, $g$ is not the uniform prior, but constructed as a mixture of MVN distributions based on a Laplace approximation, each centered at a local optimum identified via the LM algorithm to place the initial particles close to non-zero probability regions.
Using the proposal distribution $g$ rather than a uniform prior leads to the following modification of annealed intermediate distributions compared to Eq.~\ref{eq:intermediate_p}:
\begin{align}
    p_t(\boldsymbol{\theta}) \propto g(\boldsymbol{\theta})^{\lambda_t} \, p(\boldsymbol{\theta} \mid h)^{\psi_t} \, p(\mathcal{D} \mid \boldsymbol{\theta}, h)^{\phi_t} ,
    \label{eq:intermediate_p_with_g}
\end{align}
where the exponents are defined as 
\begin{align*}
    \lambda_t = \max\left(0, \frac{\gamma - \phi_t}{\gamma}\right),
    \qquad
    \psi_t = \min\left(1, \frac{\phi_t}{\gamma}\right).
\end{align*}
In this way, the proposal distribution dominates in the beginning of the optimization process, and gradually decreases in importance as the algorithm converges to the posterior.

For each particle $i$, a log-likelihood $\log p(\mathcal{D} \mid \boldsymbol{\theta}^{(0)}_{i}, h)$ is computed and an associated weight assigned, initialized uniformly as $w_{i,0} = \frac{1}{N_p}$.
A weight reflects how much of the probability mass a particle contributes to the target distribution.
Weights are updated using importance sampling according to a ratio of consecutive intermediate distributions, so that
\begin{align}
    w_{t,i} = w_{t-1,i} \cdot \frac{p_t( \theta_{t,i})}{p_{t-1}(\theta_{t,i}) }.
\label{eq:weight_update}    
\end{align} 

After the weight update, a resampling step is performed if the effective sample size, computed as $ESS = \frac{1}{\sum_i{\tilde{w}_i}^2}$ using the normalized weights $\tilde{w}_i$, is smaller than a previously defined threshold. 
This happens when the weights are highly imbalanced, with only a few particles contributing most of the probability mass while the majority are close to zero.
To avoid this particle degeneracy, particles with low weight are replaced by particles with high weight, comparable to a roulette wheel selection where the chance of selecting a particle is proportional to its weight.

Next, a few iterations of MCMC mutations are performed to guarantee diversity among particles within the high-density regions. 
The first step of the Metropolis-Hastings algorithm~(see Sec.~\ref{sec:bayes_inference}) proposes a new parameter value by sampling from the proposal function using a covariance calculated from the current particle population.
This covariance serves the purpose of an adaptive step size of the proposal: if the particles are spread apart, the new particles can be sampled from a larger region (i.e., exploration), but if they are closer together, the new particles will be generated around the current optimum (i.e., exploitation).
In the second step of the MH algorithm, the mutated value $\theta'$ is accepted with a probability $\alpha$: 
\begin{align}
    \alpha = \frac{ p_t(\boldsymbol{\theta}')}{ p_t(\boldsymbol{\theta})}.
    \label{eq:acceptance_alpha}
\end{align}

These steps of reweigh, resample, and mutate are repeated iteratively, while the step size $\Delta\phi = \phi_{t+1}-\phi_{t}$ is adapted at each iteration to meet a certain target ESS, which is set to $0.8N_p$.
The expected ESS for a proposed $\phi_{t+1}'$ is computed as
\begin{align*}
    ESS(\phi_{t+1}') &= \frac{\left(\sum_{i=1}^{N_p} \beta_i\right)^2}
    {\sum_{i=1}^{N_p}  \beta_i^2}, \text{ where }
    \beta_i = \frac{w_{i,\phi_{t+1}'}}{w_{i,\phi_{t}}}
\end{align*}

The expected weight $w_{i,\phi_{t+1}'}$ at the proposed $\phi_{t+1}'$ is calculated by inserting $\phi_{t+1}'$ into Eqs.~\ref{eq:intermediate_p_with_g} and~\ref{eq:weight_update} at $p_t(\boldsymbol{\theta})$.
A bisecting algorithm identifies the optimal $\phi_{t+1}$ by computing the expected ESS at the midpoint between $\phi_{t}$ and 1, and shrinking the interval bounds until the target ESS is achieved. 

The algorithm stops when $\phi_T=1$, resulting in the posterior distribution. 
All samples and associated weights are stored over the time steps and used to calculate the numerator and denominator of the NMLL at $\phi_t=1$ and $\phi_t= \gamma$ respectively:
\begin{align*}
    \int p(\boldsymbol{\theta} \mid h) p(\mathcal{D} \mid \boldsymbol{\theta}, h)^{\phi_t} d \boldsymbol{\theta} \approx \sum_{i=1}^{N_p} w_i(\phi_t)
\end{align*}
The resulting marginal likelihoods are inserted into the expression for $q_i(\gamma)$, see Eq.~\ref{eq:fbf}.

The admittedly complex computation of the NMLL can serve two purposes: fitting the parameters with quantified uncertainties, and providing a quality measure for symbolic models.
\citet{leser_comparing_2024} purely focus on the NMLL estimation, decoupled from the evolutionary GPSR setting, and compare three approaches: sequential Monte Carlo (SMC), Laplace approximation with direct optimization using the Levenberg–Marquardt~(LM) algorithm (L-DO), and Laplace approximation combined with SMC (L-SMC).
For unimodal posteriors, L-DO produced larger variances across repeated experiments than SMC and L-SMC, indicating a high sensitivity of the Laplace approximation to the locally estimated MAP point and curvature.
SMC-based methods captured multiple modes better in multimodal settings compared to the Laplace approximation, but showed ill-conditioned behavior for narrow modes and large distances between modes.
Due to the increased runtime of SMC and L-SMC compared to L-DO, the choice of NMLL estimator is a tradeoff between efficiency and accuracy.
An improved hyperparameter setting for SMC-based methods to reduce runtime was left for future work. 

In their most recent publication, \citet{bomarito_bayesian_2026} approximate the posterior distribution \emph{over symbolic model structures} using SMC.
While they previously focused on posterior distributions over continuous parameters $\boldsymbol{\theta}$ (parameter uncertainty)~\cite{bomarito_bayesian_2022,bomarito_automated_2023,leser_comparing_2024}, the new approach marginalizes over parameters to operate directly on the discrete hypothesis space $\mathcal{H}$.
Each symbolic model $h \in \mathcal{H}$ defines a functional mapping $f(\mathbf{x};\boldsymbol{\theta})$, and its posterior probability is determined through the marginal likelihood, which integrates out parameter uncertainty so that 
\begin{align}
    p(h \mid \mathcal{D}) &= \int_{\mathbb{R}^{d}} p(\boldsymbol{\theta}, h \mid \mathcal{D}) d\boldsymbol{\theta} \; \label{eq:marginal}\\
     &\propto \; 
p(h) \int_{\mathbb{R}^{d}} p(\mathcal{D} \mid \boldsymbol{\theta}, h)\,p(\boldsymbol{\theta} \mid h)  d\boldsymbol{\theta} \nonumber.
\end{align}
A uniform model structure prior $p(h)$ was used, avoiding prior assumptions about the frequency of functional operators, and expressions are generated at random.
The SMC-based method of likelihood tempering, as introduced in Eq.~\ref{eq:intermediate_p} for the posterior over parameters, is adapted to approximate the posterior over the model space:
\begin{align}
    p_t(h\mid \mathcal{D}) \propto p(h)\hat{q}(h)^{\phi_t}.
    \label{eq:likelihood_tempering_models}
\end{align}

Building upon the results of their previous study~\cite{leser_comparing_2024}, the NMLL $\hat{q}$ is estimated using Laplace approximation around the MAP $\boldsymbol{\theta}^*$, mainly due to the computational overhead of using an inner loop of SMC for parameter estimation within an outer loop of SMC for sampling model structures. 
The main design decisions and outer loop of the SMC algorithm remain the same: the step size $\Delta \phi_t$ for the update is adapted by a bisecting algorithm targeting a certain ESS, weights are updated using importance sampling, and the method of stratified resampling is employed to prevent particle degeneracy. 
Operating on the model space $\mathcal{H}$, weight updates are no longer based on the log-likelihood for a certain parameter value $\boldsymbol{\theta}$, but using the NMLL $\hat{q}$ as a quality measure for a symbolic model, marginalized over the space of $\boldsymbol{\Theta}$:
\begin{align*}
    w_{t,i} = w_{t-1,i} \cdot \frac{\hat{q}_t(h_{t,i})}{\hat{q}_{t-1}(h_{t,i}) }.
\end{align*} 
Resampled and thus duplicate particles (or expressions) in the population are mutated with a few MCMC moves. 
At every MCMC move, an offspring is generated from every particle in the population using standard genetic operators for crossover and mutation. 
The offspring's parameters are then refitted using the LM algorithm, the equation is evaluated and an acceptance probability $\alpha$ computed similar to the one in Eq.~\ref{eq:acceptance_alpha}, but for the proposed model $h'$ instead of parameter vector $\boldsymbol{\theta}'$.
The offspring replaces its parent with the probability $\alpha$.
These steps, from updating $\phi_t$ to performing the MCMC moves, are repeated until $\phi_t=1$.
At this point, the final population of particles approximates the posterior distribution over symbolic models $p(h \mid \mathcal{D})$, and the particles can be treated as weighted samples drawn from this posterior. 
To prevent misunderstandings, we want to clarify that sampling from the posterior does not generate new symbolic expressions, but selects expressions from the final population according to their posterior weights, i.e., only models contained in the approximated posterior can be sampled.

The SMC-based SR is compared to three evolution-based GPSR algorithms, which differ in their selection method (probabilistic crowding and tournament selection) and fitness measure (RMSE and NML).
On noisy datasets, the SMC-SR algorithm outperformed the GP variants in terms of both training and testing error and developed solutions less prone to overfitting.
The authors attributed the improved generalization capabilities of SMC-SR not only to the use of NMLL and selection pressure induced by resampling, but also to the combination of probabilistic selection (Eq.~\ref{eq:acceptance_alpha}) and likelihood tempering (Eq.~\ref{eq:likelihood_tempering_models}).
The stepwise introduction of larger portions of the likelihood, together with an acceptance probability of mutated individuals, seem to provide a structured way of balancing exploration and exploitation, which is only regulated through the selective pressure in standard GPSR. 

Overall, this body of research~\cite{bomarito_bayesian_2022,bomarito_automated_2023,leser_comparing_2024,bomarito_bayesian_2026} on SMC for SR demonstrated how to integrate Bayesian ideas into the evolutionary framework of GP on various levels, which helped to generate more accurate and generalizing equations on most datasets with noise. 
Across these four publications, the authors incrementally tackle different aspects of Bayesian approaches to SR, including model selection, NMLL estimation, posteriors over parameters and model structures, as well as highlighting the possibility of generating credible intervals, which contribute to the overall understanding of the Bayesian formulation of SR.
Particular emphasis is placed on a fair comparison between the SMC-SR and GPSR approaches, where algorithmic design decision were justified by findings of their previous studies.
The results suggest various directions for future research, such as developing more efficient methods to compute the NMLL using SMC.

\subsubsection{Bayesian SR with model structure priors}%
Guimera et al.~\cite{guimera_bayesian_2020,guimera_bayesian_2026} also employ the marginal likelihood from Eq.~\ref{eq:marginal} as a fitness measure, and marginalize over the parameters using Laplace approximation. 
The prior over model structures $p(h)$ is kept in their formulation, as this work uses an informative prior that is not assumed to be uniform:
\begin{align*}
    \log p(h \mid \mathcal{D}) = \frac{\text{BIC}(h)}{2} - \log{p(h)}
\end{align*}

To estimate $p(h)$, they calculate the average number of operators observed on a corpus of mathematical expressions extracted from the Wikipedia
\begin{align*}
    p(h) = \sum_{\operatorname{op} \in \mathcal{O}}{\left[\alpha_{\operatorname{op}} n_{\operatorname{op}}(h) + \beta_{\operatorname{op}} n_{\operatorname{op}}^2(h)\right]},
\end{align*}
where $\mathcal{O}$ is the set of operators, $n_{\operatorname{op}}(h)$ and $n_{\operatorname{op}}^2(h)$ are the average and squared frequencies of operator $\operatorname{op} \in h$, and $\alpha_{\operatorname{op}}$ and $\beta_{\operatorname{op}}$ are hyperparameters fitted so that expressions generated using this prior are consistent with the corpus.
These hyperparameters are fitted using MCMC by sampling a large number of functions $f$ from the current $p(h)$ and updating the parameter values as
\begin{align*}
    \alpha_{\operatorname{op}} &\leftarrow \alpha_{\operatorname{op}} + \epsilon \lambda \frac{\langle n_{\operatorname{op}}\rangle^{m} - \langle n_{\operatorname{op}}\rangle^{t}}{\langle n_{\operatorname{op}}\rangle^{t}} \\
    \beta_{\operatorname{op}} &\leftarrow \beta_{\operatorname{op}} + \epsilon \lambda \frac{\langle n_{\operatorname{op}}^2\rangle^{m} - \langle n_{\operatorname{op}}^2\rangle^{t}}{\langle n_{\operatorname{op}}^2\rangle^{t}},
\end{align*}
where $\langle . \rangle^m, \langle . \rangle^t$ is the mean of the mean frequencies over the sampled functions and the corpus, respectively, $\epsilon$ is a random number in $[0, 1]$ and $\lambda$ is a fixed hyperparameter.

To find expressions that fit the data, they apply another MCMC starting with a random function $f$, where one of the three moves is selected at each step with respective probabilities of $5\%, 45\%$ and $50\%$: \textbf{node replacement} replaces a random node with a random operator of the same arity; \textbf{root replacement} adds a random operator at the root of the tree or returns the left child of the current root with equal probability. When adding a root, if the chosen operator is unary, the child is the original tree, if it is binary, it will choose a random terminal to be the right child; and \textbf{tree replacement} replaces a random subtree with at most $3$ nodes from $h$ with a randomly generated tree with also at most $3$ nodes.
After applying one of these moves, the new expression will be accepted using a move-dependent MH acceptance scheme~\cite[Eq.~8-11]{guimera_bayesian_2020}.

This algorithm was capable of finding expressions balancing accuracy and complexity on a selection of artificial and real-world datasets~\cite{guimera_bayesian_2020} and it was shown to be more robust against overfitting in a large noise regime for a simple dataset~\cite{guimera_bayesian_2026}.
A limitation of this work is that the structural prior disregards the context where nodes are placed in the expression.

\subsubsection{A better structural prior for symbolic regression}%
\citet{bartlett_priors_2023} mitigate the lack of context in the model priors by modeling node probabilities in expression trees conditional on their ancestors. 

They define a tree $T = (\mathcal{L}_0, \dots, \mathcal{L}_D)$ as a collection of collections\footnote{Note: in~\cite{bartlett_priors_2023}, the authors define sets, while their implementation uses collections (as traditionally used in $n$-grams).} of nodes at every level of the tree, starting from the root at level 0 until its depth $D$.
In this context, $\mathcal{L}_0 = (r)$ contains only the root node $r$, and $\mathcal{L}_i$ contains every node at the level $i$.
The probability of producing a tree is given by $p(T) = \prod_{i=0}^{d-1}{p(\mathcal{L}_i \mid \mathcal{L}_{i-1}, \ldots, \mathcal{L}_0})$, which is estimated using a language model trained on a corpus of well established equations.
The language model estimates the probability $p(w_i \mid w_{i-1}^{i-n+1})$\footnote{For brevity, we adopt $w_i^j = (w_i, \ldots, w_j)$.} of a word $w_i$ as the relative frequency of occurrence of the sentence $w_i^{i-n+1}$ in this corpus. 
In their approach, a sentence contains $n$ words, also referred to as n-gram. 
Applying this idea to a symbolic tree, the probability of any collection of siblings only depends on their ancestors up to $n-1$ levels above.
For never observed sentences, these models return a non-zero probability using a smoothing technique.

To simplify the probability calculation, they split the collection $\mathcal{L}_i$ at level $i$ into a collection of siblings as $\mathcal{L}_i = (s_i^0, \ldots, s_i^{n_i-1})$, where $n_i$ is the number of siblings at level $i$. 
In the example $\sin(x) + \sin(x-y)$ with $+$ at the root node, the collection $\mathcal{L}$ is
\begin{align*}
    &\mathcal{L}_0 = ((+)),
    \mathcal{L}_1 = ((\sin, \sin)),
    \mathcal{L}_2 = ((x), (-)),
    \mathcal{L}_3 = ((x, y))
\end{align*}
with the corresponding sentences for $3$-grams $(+, \sin, \sin)$, $(+, \sin, x, \emptyset)$, $(+, \sin, -, \emptyset)$, $(\sin, -, x, y)$, where the sentences are padded with $\emptyset$ to maintain consistency when there is only a single sibling, i.e., input of a unary function.
Then, $p(T)$ is approximated as
\begin{align*}
    p(T) &\approx \prod_{i=0}^{d-1}{\prod_{j=0}^{n_i-1}{p(s_i^j \mid a^{ij}_{i-1}, \ldots, a^{ij}_{i-(n-1)})}}, 
\end{align*}
where $s_i^j$ is the $j$-th collection of siblings at level $i$ and $a^{ij}_{k}$ is the direct ancestor of siblings $s_i^j$ at depth $k$. 
The probability of this same example expression  would be calculated as
\begin{align*}
 p(T) &= p(+) p((\sin, \sin) | +) p(x | \sin, +)  p(- |\sin, +)  \\ & p((x,y) | -,\sin).
\end{align*}
Notice that the probability of any $s_i^j$ with two siblings is calculated as the probability of the first siblings given the $n-1$ ancestors multiplied by the probability of the second sibling given the first sibling and the $n-2$ ancestors to form the $n$-gram. 
Thus, the example becomes
\begin{align*}
 p(T) &= p(+) p(\sin | +) p(\sin | +, \sin) p(x | \sin, +)  p(- |\sin, +)  \\ & p(x | -,\sin) p(y | -,x).
\end{align*}

The probability of a word $w_i$ is estimated with the Katz back-off~\cite{katz_estimation_1987} model as:
\begin{align*}
    &p(w_i \mid w_{i-1}^{i-n+1}) = \\
&\begin{cases} 
\frac{C^*(w_i^{i-n+1})}{C(w_i^{i-n+1})} \frac{C(w_{i}^{i-n+1})}{C(w_{i-1}^{i-n+1})} & \text{if } C(w_{i}^{i-n+1}) > k\\
\alpha(w_{i-1}^{i-n+1}) p(w_i \mid w_{i-1}^{i-n+2}) & \text{otherwise} 
\end{cases},
\end{align*}
where $C(w_i^j)$ is the number of times a sentence $w_i^j$ appears in the corpus, $k$ is the minimum frequency threshold of a sentence, and $C^*$ is the Good-Turing estimate of $C$\footnote{\texttt{\url{https://github.com/DeaglanBartlett/katz/blob/main/katz/good_turing.py}}}.
The back-off weight $\alpha$ is
\begin{align*}
\alpha_{w_{i-1}^{i-n+1}} &= \frac{1 - \sum_{w_i: C(w_{i}^{i-n+1}) > k} d_{w_{i}^{i-n+1}} \frac{C(w_{i}^{i-n+1})}{C(w_{i-1}^{i-n+1})}}{\sum_{w_i: C(w_{i}^{i-n+1}) \leq k} p(w_{i} \mid w_{i-1}^{i-n+2})}. 
\end{align*}

The authors tested this prior in the task of model selection from a set of enumerated expressions.
While MDL as proposed in~\cite{bartlett_exhaustive_2024}~(see Sec.~\ref{sec:MDL_bartlett}) seems to consistently select the ground-truth, they noticed that this prior helped to remove accurate, but undesired expressions, from the higher ranked expressions (e.g., $x^x$).
This structural prior is sensitive to the corpus of equations and can prevent the exploration of novel expressions not present in it, even if they would be considered reasonable.
Since this paper focused only on the model selection task of a limited enumerated search space, additional experiments are necessary to better understand the potential benefits of this prior.

\subsubsection{Likelihood with uncertainties in the x-axis}%
Similarly to the intrinsic error of $y$, the independent single variable $x$ can also contain measurements uncertainties, which are often ignored.
\citet{bartlett_marginalised_2023} formulate the likelihood function incorporating the $x$-errors assuming the existence of a true value $x_t$.
They test different priors which enable integrating $x_t$ out, leading to a likelihood function that only depends on the observed values and the estimated variances.
When assuming $x_t \sim \mathcal{N}(\mu,w)$, the probability function of the true values is
\begin{align*}
&p\left(\boldsymbol{\theta},  x_t \mid x, y\right) = \prod_{i} \frac{1}{\sqrt{2 \pi w^{2}}} e^{-\frac{\left(x_{\mathrm{t}}^{(i)}-\mu\right)^{2}}{2 w^{2}}} \\
&\times \frac{1}{\sqrt{2 \pi \sigma_{x}^{(i) 2}}} e^{-\frac{\left(x_{\mathrm{t}}^{(i)}-x^{(i)}\right)^{2}}{2 \sigma_{x}^{(i) 2}}} 
\times \frac{1}{\sqrt{2 \pi s^{(i) 2}}} e^{-\frac{\left(f^{(i)}-y^{(i)}\right)^{2}}{2 s^{(i) 2}}},\nonumber
\end{align*}
where $x_t^{(i)}$ is the true value of the $i$-th data point, $x^{(i)}$ and $ y^{(i)}$ are the observed (noisy) values, $\sigma_x^{(i)}$ and $\sigma_y^{(i)}$ are the variances of $x$ and $y$, $f^{(i)}$ is the predicted value, $s^{(i)2} = \sigma_y^{(i)2} + \sigma_{int}^2$ and $\sigma_{int}^2$ is the variability not caused by measurement error\footnote{In the context of this publication~\cite{bartlett_marginalised_2023}, the notation $\square^{(i)}$ refers to the $i$-th data point in the dataset, and should not be confused with the notation used elsewhere in this paper, where superscripts index MCMC states.}.
Marginalizing over $x_t$ and adapting to nonlinear regression models, as introduced in~\cite{affenzeller_inefficiency_2024}, leads to 
\begin{align*}
&\ell_N\left(\boldsymbol{\theta} \mid x, y\right) \\
&=-\frac{1}{2} \sum_{i} \frac{w^{2}\left(f^{(i)}-y^{(i)}\right)^{2}+\sigma_{x}^{(i) 2}\left(f(\mu; \boldsymbol{\theta})-y^{(i)}\right)^{2}}{f'^{2} w^{2} \sigma_{x}^{(i) 2}+s^{(i) 2}\left(w^{2}+\sigma_{x}^{(i) 2}\right)} \nonumber\\
&-\frac{1}{2} \sum_{i} \frac{s^{(i) 2}\left(x^{(i)}-\mu\right)^{2}}{f'^{2} w^{2} \sigma_{x}^{(i) 2}+s^{(i) 2}\left(w^{2}+\sigma_{x}^{(i) 2}\right)} \nonumber\\
&-\frac{1}{2} \sum_{i} \log \left(f'^{2} w^{2} \sigma_{x}^{(i) 2}+s^{(i) 2}\left(w^{2}+\sigma_{x}^{(i) 2}\right)\right)  + \log{(2\pi)}, \nonumber
\end{align*}
where $f'$ is the derivative of $f$ (i.e., slope).
The term $f(\mu; \boldsymbol{\theta})$ can be estimated with a linear approximation $f(\mu;\boldsymbol{\theta}) \approx f'_i(\mu-x^{(i)}) + f^{(i)}$, leading to
\begin{align*}
    \ell_N &= -\frac{1}{2} \sum_{i=1}^{N} \left[ \log(2\pi) + \log(\Delta^{(i)}) + \frac{w^2(f^{(i)} - y^{(i)})^2}{\Delta^{(i)}}\right. \\
    &\left.+ \frac{ \sigma_{x}^{(i)2}(f'^{(i)}(\mu - x^{(i)}) + f^{(i)} - y^{(i)})^2 + s^{(i)2}(x^{(i)} - \mu)^2}{\Delta^{(i)}} \right], 
\end{align*}
where $\Delta^{(i)} = f'^{(i)2} w^2 \sigma_{x}^{(i)2} + s^{(i)2}(w^2 + \sigma_{x}^{(i)2})$. 

The only article that uses this modified likelihood function in the context of SR~\cite{affenzeller_inefficiency_2024} applies it to a challenging astrophysical dataset with the objective of assessing the efficiency of the genetic programming algorithm.
As observed in~\cite{affenzeller_inefficiency_2024}, this likelihood is computationally expensive, which can hinder the search under limited budget.
However, since error measurements in the $x$-axis are common in certain SR applications, this work provides considerable benefits for such situations.

\subsubsection{Ensemble of models with Bayesian Model Averaging} \citet{agapitos_ensemble_2014} propose the use of Bayesian Model Averaging over a population of diverse (but accurate) models to account for the approximation uncertainty and avoid overfitting.
In their approach, they define the probability of a prediction $y_i$ over $K$ models as
\begin{align}
    p(y_i \mid f_1,\ldots,f_k) = \sum_{k=1}^{K}{w_k g_k(y_i \mid f_{ki})},
    \label{eq:ensemble}
\end{align}
where $w_k$ is the probability that the prediction of $f_k$ is the best one among the $K$ models, and $g_k$ is the Gaussian likelihood centered around $f_k$ with a standard-deviation $\sigma_k$.

To estimate $w_k$ and $\sigma_k$, they use an Expectation-Maxi-mization~(EM) algorithm by introducing a latent variable $z_{ki}$ that represents the posterior probability of model $k$ for the prediction $y_i$. 
By construction, $z_{ki} = 1$ if the model $k$ is the best predictor for the $i$-th sample, and $z_{ki} = 0$ otherwise.
EM is initialized with $w_k = 1/K$ and an initial guess for $\sigma_k$ as the variance of the $k$-th model outputs, and then updates $z_{ki}, w_k^{(t)}, \sigma_k^{2(t)}$ in this order:
\begin{align*}
    &z_{ki}^{(t)} = \frac{g(y_i \mid f_{ki}, \sigma_k^{(t-1)})}{\sum_{m=1}^k{g(y_i \mid f_{mi}, \sigma_m^{(t-1)}})}, \\
    &w_k^{(t)} = \frac{1}{N}\sum_{i=1}^N{z_{ki}^{(t)}}, \qquad
    \sigma_k^{2(t)} = \frac{\sum_{i=1}^N{z_{ki}^{(t)}(y_i - f_{ki}})^2}{\sum_{i=1}^N{z_{ki}^{(t)}}}.
\end{align*}

The main algorithm is a traditional GP with fitness sharing, where the fitness of the $j$-th individual is
\begin{align*}
    \operatorname{SE}_{f_j} &= \operatorname{RMSE}\left(1 + \sum_{k=1,k\neq j}^{K} \mathcal{K}_{\lambda}(s_{f_j}, s_{f_k})\right), \\
    \text{with } &
    \mathcal{K}_{\lambda}(s_{f_1}, s_{f_2}) = \left(e^{\frac{\|s_{f_1} - s_{f_2}\|}{\lambda}} + 2 + e^{-\frac{\|s_{f_1} - s_{f_2}\|}{\lambda}}\right)^{-1},
\end{align*}
where $s_{f_i} = [f_i(x_0), \ldots, f_i(x_N)]$ is the vector of semantics of $f_i$.
When the algorithm reaches the stopping criterion, the $K$ best individuals are selected to create the ensemble in Eq.~\ref{eq:ensemble}.

They also explore an alternative ensemble where they first apply a $k$-Means clustering in the semantic space of the final population with $k = \{2,3,\ldots,36\}$.
Afterwards, for each $k$, they select the best individual of each cluster and create an ensemble as in Eq.~\ref{eq:ensemble}.
Finally, these $35$ ensembles form a final ensemble by modifying this equation to
\begin{align*}
    p(y_i \mid e_1,\ldots,e_k) = \sum_{k=1}^{K}{w_k g_k(y_i \mid e_{ki})}.
\end{align*}

While the ensemble models were consistently better than the baseline, the main appeal of this approach is the observed absence of undefined behavior in the predictions, even when extrapolating. 

\subsubsection{Linear Bayesian forest of probabilistic symbolic trees}
\label{sec:roypapers}
Roy et al.~\cite{roy_hierarchical_2025,roy_vasst_2026} provide a different formulation of Bayesian SR in two iterative publications.
Both papers have not been officially published at the time of this survey but are included due to their methodological novelty and the sometimes long review cycles in this research area.
Doubts regarding their experimental evaluation remain, and will be discussed in more detail after introducing the approaches. 

The proposed \texttt{HierBOSSS} framework~\cite{roy_hierarchical_2025} is a forest of $K$ linearly combined symbolic trees to predict a target variable
\begin{align*}
    y_i = \beta_0 + \sum_{j=1}^{K} \beta_j \, g(\mathbf{x}_i;\mathcal{T}_j) + \epsilon_i,
\end{align*}
where $g(\mathbf{x};\mathcal{T}_j)$ corresponds to the output of a single symbolic tree $\mathcal{T}_j$.
This linear combination of SR trees is motivated by the conjugate Bayesian inference, allowing closed-form evaluations of the marginal likelihood, as we have briefly discussed in Sec.~\ref{sec:credible_int}. 
The complete Bayesian hierarchical form is then given by
\begin{align*}
    \mathbf{y} = \mathbb{T}(\mathbf{X})\boldsymbol{\beta} + \boldsymbol{\varepsilon},
\end{align*}
with normally distributed noise ${\boldsymbol{\varepsilon}} \sim \mathcal{N}_N(0_N, \sigma^2 I_N)$ for $N$ samples in the dataset. 
The prior of the linear coefficients jointly with the error variance follow a Normal Inverse-Gamma prior, i.e., $(\boldsymbol{\beta}, \sigma^2) \sim \text{NIG}(\nu, \lambda, \mu_{\boldsymbol{\beta}}, \Sigma_{\boldsymbol{\beta}})$.
The design matrix $\mathbb{T} = (\mathcal{1}_N,g(\mathbf{x};\mathcal{T}_1), \dots, g(\mathbf{x};\mathcal{T}_K))$ has the shape $N \times (K+1)$ and contains the outputs of the symbolic trees. 
In this setting, due to conjugacy between the Gaussian likelihood and the NIG prior, the posterior over the linear regression parameters remains in the same family
\begin{align*}
(\boldsymbol{\beta}, \sigma^2 \mid \mathbb{T}, \mathbf{y}) \sim \text{NIG}(\mu_n, \Sigma_n, a_n, b_n).
\end{align*}

Since the symbolic trees enter the model only through the design matrix $\mathbb{T}$, conjugacy is required only between the Gaussian likelihood and the NIG prior on the regression parameters, while the symbolic tree priors can be non-conjugate.
The symbolic tree priors in \cite{roy_hierarchical_2025} are defined for three elements: 
the probability that a node is non-terminal, the probabilistic assignment of an operation at non-terminal nodes, and the probabilistic assignment of features at terminal nodes. 
Every node at depth $m$ is a non-terminal node with a probability $p_m = \alpha(1+m)^{-\delta_0} $, with user-adjusted parameters $\alpha$ and $\delta_0$. 
The weights for $n_o$ operations and $n_x$ features follow Dirichlet distributions
\begin{align*}
    w_{\mathrm{op},j} \sim \mathrm{Dir}(\alpha_{\mathrm{op},1}, \ldots, \alpha_{\mathrm{op},{n_o}}),
    w_{\mathrm{ft},j} \sim \mathrm{Dir}(\alpha_{\mathrm{ft},1}, \ldots, \alpha_{\mathrm{ft},{n_x}}),
\end{align*}
which parameterize categorical distributions to assign operators at non-terminal nodes and features at terminal nodes.
The prior over a symbolic tree $\mathcal{T}_j$ is thus given by
\begin{align*}
&p(\mathcal{T}_j \mid w_{\mathrm{op},j}, w_{\mathrm{ft},j}, \alpha, \delta_0) = \\
&\prod_{o=1}^{{n_o}} (w_{\mathrm{op},j,o})^{\xi_{j,o}} 
\prod_{h=1}^{n_x} (w_{\mathrm{ft},j,h})^{\varrho_{j,h}} \prod_{m=1}^{\infty} p_m^{\mathfrak{N}_{j,m}} (1 - p_m)^{\mathfrak{T}_{j,m}},
\end{align*}
with $\mathfrak{N}_{j,m}, \mathfrak{T}_{j,m}$ corresponding to the number of non-terminal and terminal nodes, respectively, in $\mathcal{T}_j$ at depth $m$, and $\xi_{j,o}$ and $\varrho_{j,h}$ denoting the frequency of operators and features in the tree. 
The joint posterior over $K$ trees in the ensemble is 
\begin{align*}
&p((\mathcal{T}_j, w_{\mathrm{op},j}, w_{\mathrm{ft},j})_{j=1}^K, \boldsymbol{\beta}, \sigma^2 \mid \mathcal{D}) 
\; \propto \;
p(\mathcal{D} \mid \mathbb{T}(\mathbf{X}), \boldsymbol{\beta}, \sigma^2) \times\\
&p(\boldsymbol{\beta} \mid \sigma^2)
p(\sigma^2) 
\prod_{j=1}^{K}
p(T_j \mid w_{\mathrm{op},j}, w_{\mathrm{ft},j}, \alpha, \delta_0)
p(w_{\mathrm{op},j})
p(w_{\mathrm{ft},j}).
\end{align*}

The posterior over a symbolic tree is non-analytical due to the discrete and combinatorial tree structure and is therefore approximated using Metropolis-within-Gibbs MCMC (as briefly introduced in Sec.~\ref{sec:bayes_inference}).
Tree structures are updated using MH proposals, where either a grow or a prune move is performed to create offspring. 
Grow replaces a randomly selected terminal node with a subtree generated from the tree prior, and prune collapses a randomly selected non-terminal node into a terminal with a feature sampled from the prior.
When updating individual symbolic trees, the joint marginal posterior~(JMP) over the full forest is evaluated for the current and proposed tree configurations and used to compute the acceptance probability.
The JMP for a single tree conditional on the rest of the forest is given by
\begin{align*}
p(\mathcal{T}_j \mid \mathcal{D}, \mathcal{F}_{-j})
&\propto \left[ \int p(y \mid \mathbb{T}(\mathbf{X}), \boldsymbol{\beta}, \sigma^2) p(\boldsymbol{\beta}, \sigma^2) \, d\boldsymbol{\beta} \, d\sigma^2 \right] \times \\
&p(\mathcal{T}_j \mid w_{\mathrm{op},j}, w_{\mathrm{ft},j}, \alpha, \delta_0),
\end{align*}
where $\mathcal{F}_{-j}$ denotes the forest without the $j^{th}$ tree that is currently being mutated. 
Due to conjugacy, the marginal likelihood integral can be computed analytically in closed form. 
This marginalization removes the sampling of regression parameters during the tree updates, and reduced the MH step to only sampling over symbolic tree structures.
After the forest update, the regression parameters ($\boldsymbol{\beta}$ and $\sigma^2$) and operator and feature weights ($w_{\textrm{op,}j}$ and $w_{\textrm{ft,}j}$) are sampled from their conjugate posteriors, which are conditional on the updated tree structures.
These steps of forest update, weights update and parameter update are repeated until a certain number of MCMC iterations is reached.
A single forest with the largest JMP is selected from the posterior.

In their follow-up publication, \citet{roy_vasst_2026} introduce \texttt{VaSST}, where they retain the linearly combined forest of symbolic trees, but replace the discrete tree search with soft symbolic trees and perform \textbf{variational inference} to approximate the posterior. 
Rather than calculating $p(f \mid \mathcal{D})$ directly, variational inference introduces a simpler family of distributions $q(f)$, whose parameters are optimized to approximate the posterior by minimizing the Kullback-Leibler~(KL) divergence
\begin{align}
    KL(q(f) \mid\mid p(f \mid \mathcal{D})) = \mathbb{E}_q[\log q(f) - \log p(f \mid \mathcal{D})]. \label{eq:kl}
\end{align}

By substituting $p(f \mid \mathcal{D}) = \frac{p(\mathcal{D}, f)}{ p(\mathcal{D})}$ in Eq.~\ref{eq:kl}
, we can rewrite the KL divergence as
\begin{align*}
\mathrm{KL}&\big(q(f) \,\|\, p(f \mid \mathcal{D})\big) = \mathbb{E}_q[\log q(f)] - \mathbb{E}_q\left[\log p(f \mid \mathcal{D})\right] \\
&=  \mathbb{E}_q[\log q(f)] - \mathbb{E}_q\left[\log \left(\frac{p(\mathcal{D}, f)}{ p(D)}\right)\right]\\
&= \mathbb{E}_q[\log q(f)] - \mathbb{E}_q[\log p(D, f)]  +\mathbb{E}_q[\log p(D)]  \\
&= \log p(\mathcal{D}) -
\underbrace{\left( \mathbb{E}_q[\log p(\mathcal{D}, f)]
- \mathbb{E}_q[\log q(f)] \right)}_{\text{ELBO}},
\end{align*}
where the bracketed term is the Evidence Lower Bound (ELBO)~\cite[Sec.~2.2]{kingma_introduction_2019}. Since the KL divergence is non-negative and the evidence $p(\mathcal{D})$ is constant with respect to $q$, minimizing the KL divergence is equivalent to maximizing the ELBO.
Applying the product rule $p(\mathcal{D}, f) = p(\mathcal{D} \mid f)p(f)$, the ELBO can be rewritten as
\begin{align*}
    \operatorname{ELBO} &= \mathbb{E}_q[\log p(\mathcal{D}, f)] - \mathbb{E}_q[\log q(f)] \\
    &= \mathbb{E}_q[\log p(\mathcal{D} \mid f) + \log p(f)] - \mathbb{E}_q[\log q(f)]  \\ 
    &= \mathbb{E}_q[\log p(\mathcal{D} \mid f)] - \mathbb{E}_q[\log q(f) - \log p(f)]  \\ 
    &= \mathbb{E}_q[\log p(\mathcal{D} \mid f)] - \operatorname{KL}\left(q(f) \mid\mid p(f)\right).
\end{align*}
where the first term represents the expected log-likelihood of the model and the second term is the KL divergence of the tractable family of functions $q(f)$ from the prior $p(f)$, which acts as a complexity penalty or regularizer~\cite[Sec.~2.2]{kingma_introduction_2019},~\cite[Sec.~10.1.1.2]{murphy_probabilistic_2023}.

The overall idea of \texttt{VaSST} is to optimize ELBO in order to achieve optimal parameters of the tree prior such that the likelihood of a sampled tree increases while the overall structure remains simple.
The optimization employs an annealing schedule for a temperature hyper-parameter $\tau$.
Starting with a high temperature facilitates broad exploration of the model structure space by allowing a uniform mixture of operators and features. 
Gradually decreasing $\tau$ toward zero forces the optimization to choose the preferred operators and features of each node, thus enabling exploitation.
For each $\tau$, the algorithm calculate the two terms, i.e., KL divergence and expected log-likelihood, of the ELBO separately.
Afterwards, the gradient is calculate and the parameter values are updated.
To enable gradient-based optimization of the ELBO, the discrete model structure variables are relaxed into soft trees.

In~\cite{roy_vasst_2026}, the prior $p(f)$ is formulated as the prior of a forest of $K$ trees, denoted by $p(\Xi)$:
\begin{align*}
    p(\Xi) = 
    p(\mathbf{w}_{\text{op}}) p(\mathbf{w}_{\text{ft}}) \prod_{j=1}^{K} \prod_{\zeta \in \mathscr{Z}_D} p(e_{j\zeta}) p(o_{j\zeta} \mathord{\mid} \mathbf{w}_{\text{op}}) p(h_{j\zeta} \mathord{\mid} \mathbf{w}_{\text{ft}}),
\end{align*}
where $\mathbf{w}_{\text{op}}, \mathbf{w}_{\text{ft}}$ are vectors representing the global preference for each operator and feature, respectively. 
The binary variable $e_{j\zeta}$ indicates whether the $\zeta$-th node of the $j$-th tree is an internal node ($e_{j\zeta} = 1$) or a terminal node ($e_{j\zeta} = 0$).
Finally, $o_{j\zeta}$ and $h_{j\zeta}$ are categorical variables specifying the particular operator or feature assigned to that node.
Each tree in the forest is defined as a full binary tree a of fixed depth $D$, so that all trees in the forest have the same number of trainable parameters. 
Consequently, $\mathscr{Z}_D$ represents the complete set of nodes in such a tree, where $|\mathscr{Z}_D| = 2^{D+1} - 1$. 

This prior cannot be calculated analytically due to the dependency of $o_{j\zeta}$ and $h_{j\zeta}$ on the global weights, which induces intractable conditional distributions over latent variables.
Thus, the formulation is simplified using a mean-field factorization, which assumes that all variables are mutually independent.
The joint variational distribution can then be reformulated as
\begin{align*}
    q_\phi(\Xi) = 
    q_\phi(\mathbf{w}_{\textrm{op}}) q_\phi(\mathbf{w}_{\textrm{ft}})
    \prod_{j=1}^{K} \prod_{\zeta \in \mathscr{Z}_D} q_\phi(e_{j\zeta}) q_\phi(o_{j\zeta}) q_\phi(h_{j\zeta}),
\end{align*}
where $\phi$ represents the set of variational parameters ($\mathbf{w}_{\textrm{op}}, \mathbf{w}_{\textrm{ft}}, e_{j\zeta}, o_{j\zeta}, h_{j\zeta}$) being optimized. 
To enable gradient-based optimization, the Gumbel-Softmax relaxation is applied to $o_{j\zeta}$, $h_{j\zeta}$, and $e_{j\zeta}$, as for example
\begin{align}
\tilde{o}_{j\zeta, k} = \frac{\exp\left( \frac{\log(\pi^{(o)}_{j\zeta, k}) + g_{j\zeta, k}}{\tau} \right)}{\sum_{c=1}^{C_o} \exp\left( \frac{\log(\pi^{(o)}_{j\zeta, c}) + g_{j\zeta, c}}{\tau} \right)}, \label{eq:softmax}
\end{align}
where $\tilde{o}_{j\zeta, k}$ represents the $k$-th element of the relaxed variable $\tilde{\mathbf{o}}_{j\zeta}$ at node $\zeta$ of tree $j$, and ${\pi}^{(o)}_{j\zeta,k}$ is the probability (or preference) of choosing the category $k$ ($\sum_k{{\pi}^{(o)}_{j\zeta,k}} = 1$). 
The temperature parameter $\tau$ controls how peaked the distributions over the relaxed variables become. 

This relaxation enables the analytical calculation of the KL term of the ELBO, as derived by the authors in~\cite[Appendix G]{roy_vasst_2026}.
Since the expectation of the log-likelihood cannot be derived analytically, it is estimated via Monte Carlo integration. 
Specifically, $S$ soft forests are generated by sampling random values $g_{j\zeta, k}$ from a standard Gumbel distribution to compute the relaxed variables $\tilde{e}_{j\zeta}$, $\tilde{\mathbf{o}}_{j\zeta}$, and $\tilde{\mathbf{h}}_{j\zeta}$\footnote{This procedure turns the random categorical variables $o_{j\zeta}$ and $h_{j\zeta}$ into vectors of probabilities for each category, therefore we apply bold notation for $\tilde{\mathbf{o}}_{j\zeta}$ and $\tilde{\mathbf{h}}_{j\zeta}$.} via Eq.~\ref{eq:softmax}.
These relaxed variables represent the preference weights for node types (internal vs. terminal), operators and features. 
The output of each soft tree is computed by propagating these evaluations upward from the leaves to the root. 
For a node $\zeta$ in the $j$-th tree, the soft evaluation $y_{j\zeta}$ is defined as
\begin{align*}
y_{j\zeta} &= (1 - \tilde{e}_{j\zeta}) \operatorname{term}_{j\zeta} + \tilde{e}_{j\zeta} \operatorname{nonterm}_{j\zeta} \\
\operatorname{term}_{j\zeta} &= \sum_{t \in \mathcal{T}} \tilde{h}_{j\zeta, t} x_t \\
\operatorname{nonterm}_{j\zeta} &= \sum_{\operatorname{op} \in \mathcal{O}} \tilde{o}_{j\zeta, o} \cdot \operatorname{op}(y_{j\zeta_l}, y_{j\zeta_r}),
\end{align*}
where $x_t$ represents the $t$-th feature, $\mathcal{T}$ is the set of terminals, and $\mathcal{O}$ is the set of non-terminal operators. 
Here, $y_{j\zeta_l}$ and $y_{j\zeta_r}$ are the soft evaluations of the left and right children, respectively. 
When $\operatorname{op}$ is a unary operator, only the left child is used as input.
The estimated ELBO $\widehat{ \mathscr{L} }(\phi)$ at each iteration of the optimization process is then calculated as
\begin{align*}
    \widehat{ \mathscr{L} }(\phi) =  \frac{1}{S}   \sum_{s=1}^{S} \log p\left( y \mid \mathbf{T}_{\text{soft}}^{(s)} \right)   -\mathrm{KL} \left( q_{\phi}(\Xi)  \,\|\, p(\Xi) \right).
\end{align*}

The parameter values are updated by calculating the gradient of $\widehat{ \mathscr{L} }(\phi)$ with respect to $\phi$ using backpropagation.

The algorithm returns the optimized $\phi^*$ with prior distributions biased towards the forests that maximize the ELBO.
Thus, sampling from this distribution is expected to produce a forest balancing model accuracy and parsimony. 
Every node of every tree in this skeleton has its own set of optimized parameter values for the independent distributions of $e_{j\zeta}$, $o_{j\zeta}$ and $h_{j\zeta}$. 
Instead of the soft evaluation of each node, a specific sample is drawn first from $e_{j\zeta}$ to determine its terminal or non-terminal property, and then from $o_{j\zeta}$ or $h_{j\zeta}$ to sample the corresponding operator or feature.
Nodes at depth $D$ are enforced to be terminals.
The process is repeated $K$ times, generating a sampled forest where the coefficients $\boldsymbol{\beta}, \sigma^2$ can then be estimated based on the data.
Sampling multiple forests delivers the estimate of the posterior distribution $p(\Xi,\boldsymbol{\beta}, \sigma^2)$.
Even though the authors mention the possibility of UQ, this is not properly described in their paper.

Both works~\cite{roy_hierarchical_2025,roy_vasst_2026} introduce promising concepts for Bayesian SR, including structural priors, exploitation of conjugacy, as well as the introduction of soft trees. 
However, a few limitations hinder the applicability for real-world data.
First, both approaches are restricted to linear combinations of trees, which limited their ability to model phenomena which are nonlinear in their parameters.
Second, the variational distribution $q$ requires a large set of latent variables to determine node types and their corresponding assignments, which increases exponentially with the depth of the tree, making variational inference feasible only for shallow structures.
The introduced prior over the model structures assumes node independence, unlike the prior introduced in~\cite{bartlett_priors_2023}, which can potentially steer the estimated posterior away from the true one (i.e., a sample of $N$ forests from this posterior may fail to concentrate in regions of high mass of the true posterior).
Finally, an investigation of the source-codes\footnote{\texttt{\url{https://github.com/Roy-SR-007/HierBOSSS/}}}$^,$\footnote{\href{https://anonymous.4open.science/r/VaSST-62C7/example.ipynb}{\texttt{https://anonymous.4open.science/r/example.ipynb}}} revealed inconsistencies with the reported experiments.
For instance, the used operator set in \texttt{HierBOSSS}\footnote{\href{https://github.com/Roy-SR-007/HierBOSSS/blob/c9c72931e18ce0cb791c42ff6a6b214dcab28d75/HierBOSSS_simulated_example.R}{\texttt{https://github.com/HierBOSSS\_simulated\_example.R}}} did not correspond to those reported in the paper~\cite{roy_hierarchical_2025}, and certain operators reported in \texttt{VaSST}~\cite{roy_vasst_2026} (exp, $\log$) were not currently implemented\footnote{\href{https://anonymous.4open.science/r/VaSST-62C7/VaSST.py}{\texttt{https://anonymous.4open.science/r/VaSST.py}}}. 
After correcting these inconsistencies and attempting to replicate the experiments, we found that the returned models failed to recover the ground-truth. 
Overall, the proposed methods represent a previously unexplored direction in SR and therefore relevant for our survey, although we believe that methodological adaptations are required to become a viable direction of future research.

\subsubsection{Quantifying uncertainty post-hoc with MCMC} \citet{zhao_uncertainty_2025} apply MCMC at the final model found by an SR algorithm for the joint rock roughness coefficient to calculate the credible interval of the coefficients and of the predictions.

\subsection{Non-Bayesian methods}
\subsubsection{Predicting conditional quantiles with SR} \citet{hoekstra_symbolic_2025} replace the fitness function of an SR algorithm to estimate the conditional quantile function for a level $\tau \in (0, 1)$:
\begin{align*}
    Q_{y}(\tau \mid x) = \inf\left\{q \in \mathbb{R} : p(y \leq q \mid x) \geq \tau \right\}.
\end{align*}

This corresponds to the minimum value $q$ in which the probability of the prediction being less than $q$ is greater than $\tau$.
Whenever this objective-function is used to generate two models, one with $\tau = \alpha/2$ and another with $\tau = 1 - \alpha/2$, the predictions of both models correspond to the prediction interval with a significance level of $\alpha$.
To find quantile regression models with SR, the authors use the pinball loss as the fitness function:
\begin{align*}
    L_{\tau}(y_{i}, f_i)=\begin{cases}(\tau-1)(y_{i} - f_i)&if~y_{i} - f_i\ge0,\\ \tau(y_{i} - f_i)&if~y_{i} - f_i<0\end{cases}.
\end{align*}
The quantile regression minimizes the effective risk of predictions outside the confidence region, as it generates a model that tries to predict a value within a range with a certainty level $\tau$. 
While there is no guarantee of achieving this level, it will guide the model toward a robust region.
The results show that the fitness function helped to achieve consistently better results than traditional quantile methods on a large selection of datasets, although the authors have not compared their results against state-of-the-art SR methods.
Even though quantile regression is capable of producing the prediction intervals without additional calculations, the authors have not explored this venue in this work.

\begin{figure}
\centering

\begin{tikzpicture}
\begin{axis}[
    width=\linewidth,
    height=0.63\linewidth,
    xlabel={$\theta_0$},
    ylabel={$\theta_1$},
    grid=both,
    xmin=14,
    xmax=33,
    ymin=-1.65,
    ymax=-0.1,
]
\addplot[
    thick,
    blue,
    smooth,
] coordinates {
(15.300961858252974,-1.1673908207046657)
(15.654915996085382,-1.2757029238727151)
(16.067738966326377,-1.3768333862582252)
(16.54037026237116,-1.4628014894842285)
(17.07483753038291,-1.5247072814714788)
(17.674722912023167,-1.554780741845425)
(18.34584994140847,-1.5487166518290074)
(19.09638933261665,-1.507409429912213)
(19.93753703815265,-1.436553412138385)
(20.88312183023784,-1.3446011117551635)
(21.948240628913076,-1.2402820306332927)
(23.14700989092608,-1.1310581727017792)
(24.488595883694295,-1.0224147351577055)
(25.969698757665526,-0.9179948528812414)
(27.56599245376176,-0.8199842161142745)
(29.19828969852775,-0.7295153473880224)
(30.504452867406865,-0.6469916794748848)
(31.432620143838708,-0.5723836483955133)
(32.027923590140844,-0.5054018267483411)
(32.32191228588264,-0.44562495012126047)
(32.33077689418235,-0.39256322208791194)
(32.05459037678614,-0.34576344354731003)
(31.476739075132702,-0.304826742799734)
(30.56476501837181,-0.26939755147088607)
(29.27196980181385,-0.2391825744824958)
(27.63415682115268,-0.2139533474821954)
(26.02013797114846,-0.19353568480293473)
(24.51981586095736,-0.1778030690759411)
(23.159673271537713,-0.16666642765875422)
(21.944253377793224,-0.16006591828233513)
(20.864619264852095,-0.1579589882718579)
(19.90673248503937,-0.16032608411253724)
(19.055173430815515,-0.16715918788417172)
(18.29612146120546,-0.17847537467469635)
(17.618293731625833,-0.19431132204652535)
(17.01349638479153,-0.21473732730403816)
(16.4761305097033,-0.23986238830348464)
(16.002711263063517,-0.26984296313875517)
(15.591344272118437,-0.3048934744785194)
(15.24125296318162,-0.34529063869149934)
(14.952319742107518,-0.39137538984111975)
(14.724820270400432,-0.44354838664962665)
(14.634218485272944,-0.47205616266497696)
(14.499556211404485,-0.5342124050004099)
(14.427144182358136,-0.6036000922940061)
(14.417053378609813,-0.6805996793714686)
(14.469116433226503,-0.7654354629364958)
(14.582919590378937,-0.857998927645795)
(14.757919329246356,-0.9576640093362815)
(14.993473244884136,-1.0629962887481228)
(15.28897069712151,-1.1713856879177502)
};
\end{axis}
\end{tikzpicture}
\label{fig:contourCI}
\vspace{-0.3cm}
\caption{Example of a confidence region of $\theta_0$ and $\theta_1$ using profile likelihood as introduced by~\cite{de_franca_prediction_2022}. This plot shows that the interval for $\theta_1$ is wider for smaller values of $\theta_0$ and tighter for larger values.}
\end{figure}
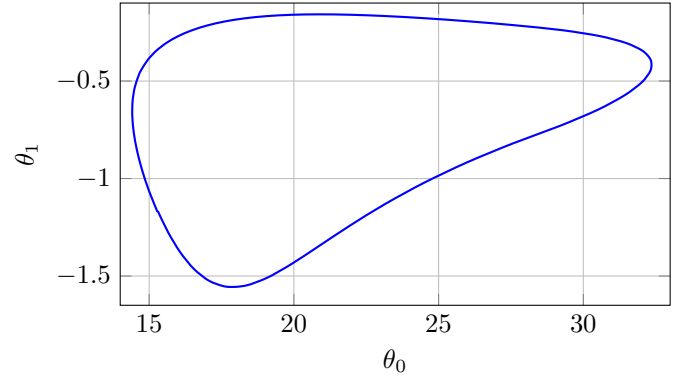

\subsubsection{Profiling the likelihood for asymmetric and continuous intervals} \citet{de_franca_prediction_2022} demonstrate the benefits of using profile likelihood to calculate the CIs of parameters and predictions in SR.
After obtaining the fitted $\boldsymbol{\hat{\theta}}$, this method creates a profiling function $\tau_i(\theta_i)$ of any arbitrary value $\theta_i$ of the $i$-th parameter as
\begin{align*}
    \tau(\theta_i) = \operatorname{sign}(\theta_i - \hat{\theta}_i) \sqrt{2\ell_N(\tilde{\boldsymbol{\theta}}) - 2\ell_N(\mathbf{\hat{\boldsymbol{\theta}}}) },
\label{eq:tau}
\end{align*}
where $\boldsymbol{\tilde{\theta}}$ are the optimal parameters values when fixing the value of $\theta_i$ and maximizing the log-likeli-hood function for $\boldsymbol{\theta}_{j\neq i}$. 
The CI at a level $\alpha$ is defined as the infimum and supremum of the set $\left\{\theta_i \mid -t(n - p, \alpha/2) \leq \tau(\theta_i) \leq t(n - p, \alpha/2)\right\}$.
If the true closed-form of $\tau$ is known, this is the exact confidence interval at that level.
However, since finding this closed-form is intractable, the function is estimated by sampling values of $\theta_i$ and applying an interpolation method, such as cubic splines.
The main advantages of this approach are: it captures the asymmetry of the intervals, it is applicable with any likelihood function, it can be use to calculate the dependence in the intervals of pairs of parameters (see Fig.~\ref{fig:contourCI}), and it can reveal issues with the current model when the intervals are unbounded.
Another contribution of~\cite{de_franca_prediction_2022} is the introduction of an algebraic manipulation algorithm which enables this method to calculate the CI of the predictions. 
While being more accurate than the delta method, this approach has a higher computational cost. 
 
\subsubsection{Structural and parametric UQ with bootstrapping} \citet{manzi_discovering_2020} apply SR to uncover the state function describing the space dynamics, whose measurements typically contain high levels of noises. 
They applied a bootstrapping technique, which generates multiple samples from the original dataset and fits each separately using an SR algorithm with parameter fitting.
Afterwards, they identified similar model structures across different bootstrap samples, and plotted the respective parameter values as histograms. 
In this way, it becomes visually apparent how much the parameter values are spread out for different noise levels, and thus how ``certain'' the parameter estimates are. 
However, no estimation of the parameter distributions was performed, which limits the findings to visual interpretation. 
Nevertheless, this representation enables informed decisions about function structures. 

\subsubsection{Terminal symbols to model noise}
\citet{schmidt_learning_2007} argue
that while handling the noise through an expectation function can suffice, in some datasets the noise can have a nonlinear relationship with the dependent variable.
This relationship can be invaluable to better understand the phenomena and to achieve the correct model structure.
For example, if the noise grows quadratically with the observations, this relationship should be contained in the prediction model.
To account for nonlinear noise relationships, they introduce a terminal symbol $R$ that is a uniformly distributed random variable $U[-1, 1]$.
This uniform distribution is capable of modeling different noise distributions, including the Gaussian.
They modify the traditional fitness function to consider the range of the prediction interval
\begin{align*}
    \text{fitness}(f) = \sum_{i=0}^n{\begin{cases}\min_{r_i \in r_{f(x_i)}}{|y_i = r_i|} & \text{if } y_i \notin r_{f(x_i)} \\ \frac{1}{r_{f(x_i)}} & \text{otherwise} \end{cases}},
\end{align*}
where $r_{f(x_i)}$ is the range of the evaluation of $f$ at $x_i$ calculated as the minimum and maximum value of $100$ distinct samples for the $R$ terminals in the expression.
The first part of the fitness function aims to minimize the distance between the observed value and the prediction range, when the observation is outside the range.
The second part aims to avoid the degenerate case where the range is wide enough to include every observation in the dataset.
With this method, the prediction intervals can be estimated by generating multiple realizations of the prediction for every predicted point.
However, this procedure does not provide the formal guarantees as in the traditional prediction intervals, as described in Sec.~\ref{sec:ci-pi}.

\subsubsection{Minimizing prediction interval with conformal prediction} \label{sec:confpred_review}
\citet{thuong_combining_2017} propose two ways of integrating conformal prediction
in SR: i) generating prediction intervals by applying CP in the best model found by SR or, ii) applying CP to every individual and incorporating the interval width in the fitness function using the weighted average $\alpha \text{RMSE}(f) + (1 - \alpha) \text{PI}_w$, where $\text{PI}_w$ is the width of the prediction intervals generated by CP.
Their experiments showed that applying CP to the final model produced better results than incorporating the prediction width into the fitness function.
Additionally, this version was capable of obtaining better results than the traditional quantile regression algorithms.
The authors comment that further investigation is necessary to understand why considering the interval width did not improve the results.

\subsection{Model selection methods}

\subsubsection{Description length for polynomial models}
\citet{iba_genetic_1994} apply the MDL principle in SR with the introduced STructured Representation On Genetic Algorithms for NOn-linear Function Fitting (STROGANOFF), where each leaf node is a variable and the intermediate nodes are the function $G(z_i, z_j) = a_0 + a_1 z_i + a_2 z_j + a_3 z_i z_j + a_4 z_i^2 + a_5 z_j^2$.
Thus, this representation produces a polynomial regression model of degree $k$.
They followed the DL general formula (Eq.~\ref{eq:MDL}) $L(\mathcal{D}) = L(h) + L(\mathcal{D} \mid h)$, with $L(\mathcal{D} \mid h) = 0.5 N \log \text{MSE}$ an $L(h) = 0.5 d \log N$, where $d$ is the total number of parameters $a_i$ and $N$ the number of data points.
They noticed through a series of experiments that using their proposed DL as a fitness, not only managed to control the expression sizes but also increased the accuracy of the final expressions.
\citet{nikolaev_regularization_2001} extend the definition of their DL as $MSE + \frac{d}{N}s^2\log N$, where $s^2$ is an estimation of the variance given by $\frac{1}{N}\sum_{i=1}^N{(y_i - \bar{y})^2}$.

\subsubsection{Description length considering parametric nonlinear models} \label{sec:MDL_bartlett}
\citet{bartlett_exhaustive_2024} propose an exhaustive search for SR (ESR) that avoids the generation of redundant expressions. 
The enumerated space requires to rank and select the best models.
For this purpose, the authors propose a DL for SR that considers multiple aspects of a model. 
As is standard in MDL literature for regression, $L(\mathcal{D} \mid h)$ is the negative log-likelihood, while the DL of the model~$L(h)$ is split into a functional and a parameter part. 
For the functional part, a tree with $k$ nodes and $n$ unique operators must transmit $k \log n$ nats of information\footnote{Analogue to a binary encoding, $n$ symbols are described with $\lg n$ bits.}.
Constant values $c$, such as natural numbers in exponents, require $\log c$ nats of information.
The $d$ real-valued parameters of the expression, in theory, require an infinite number of nats of information to be precisely encoded.
The authors discretize the values with a grid of fixed width $\Delta_i$.
For example, a $\Delta_i = 0.2$ would describe the values $\ldots,-0.2,0,0.2,0.4,\ldots$, and any number between two bins will be rounded to the nearest value.
Therefore, the optimal parameter with infinite precision will fall uniformly at random between $[\hat{\theta}_i - \Delta_i/2, \hat{\theta}_i + \Delta_i / 2]$.
The codelength of the parameters is then $\log (|\theta_i|/\Delta_i) + \log 2$ nats, where $1/\Delta_i$ is the precision and $\log 2$ encodes the sign.
A large value of $\Delta_i$ would decrease the parameter codelength but would potentially increase the log-likelihood, as the parameter value would be imprecise.
Applying the Taylor expansion on $-\ell_N(\hat{\boldsymbol{\theta}}+\delta)$, where $\delta$ is a uniform random variable in $[- \Delta_i/2, \Delta_i/2]$, leads to
\begin{align*}
    \ell_N(\hat{\boldsymbol{\theta}}+\delta) &\approx \ell_N(\hat{\boldsymbol{\theta}}) + \delta \, \ell'_N(\hat{\boldsymbol{\theta}}) + \frac{\delta^2}{2}\ell''_N(\hat{\boldsymbol{\theta}}) \\ 
    &\approx \ell_N(\hat{\boldsymbol{\theta}}) + \frac{1}{2}\delta^TI(\boldsymbol{\theta})\delta,
\end{align*}
since $\ell'_N(\hat{\boldsymbol{\theta}}) = 0$ according to the optimality criteria, and $I$ is the Fisher information matrix.
The contribution of $\Delta_i$ to the codelength becomes
\begin{align*}
    L(\boldsymbol{\Delta}) = \frac{1}{2}\sum_{ij}{I_{ij}\mathbb{E}[\delta_i\delta_j]} - \sum_i{\log \Delta_i}.
\end{align*}
The authors assume that the random variables $\delta_i, \delta_j$ for $i\neq j$ are independent from each other, leading to $\mathbb{E}[\delta_i \delta_j] = \mathbb{E}[\delta_i]\mathbb{E}[\delta_j] = 0$.
Additionally following the uniform distribution, the variance of each variable is $\mathbb{E}[\delta_i^2] = \Delta_i^2/12$.
The codelength for a given $\Delta_i$ now becomes
\begin{align*}
    L(\Delta_i) = \frac{1}{24} I_{ii} \Delta_i^2 - \log \Delta_i,
\end{align*}
with a minimum at $\Delta_i = \sqrt{12/I_{ii}}$. 
Adding all these codelength components results in the definition of DL as
\begingroup
\allowdisplaybreaks
\begin{align*}
    L(D) &= -\ell_N(\hat{\boldsymbol{\theta}}) + k \log n + \sum_j{\log c_j} \\ & + \sum_{i=1}^{d}{\left(\log{(|\theta_i|/\Delta_i)} + \log{2}\right)}\\ 
    &= -\ell_N(\hat{\boldsymbol{\theta}}) + k \log n + \sum_j{\log c_j} \\&+ d (\log2 - \frac{1}{2} \log 12) + \sum_{i=1}^d{\left(\frac{1}{2}\log I_{ii} + \log |\hat{\theta}_i|\right)} \nonumber \\
    &= -\ell_N(\hat{\boldsymbol{\theta}}) + k \log n + \sum_j{\log c_j} \nonumber \\ &+ d \left(\log2 - \frac{1}{2} (\log 3 + \log 4)\right) \nonumber \\&+ \sum_{i=1}^d{\left(\frac{1}{2}\log I_{ii}  + \log |\hat{\theta}_i|\right)} \nonumber \\
    &= -\ell_N(\hat{\boldsymbol{\theta}}) + k \log n + \sum_j{\log c_j} \nonumber \\&+ d \left(\log2 - \frac{1}{2} (\log 3 + 2 \log 2)\right) \nonumber \\ &+ \sum_{i=1}^d{\left(\frac{1}{2}\log I_{ii}  + \log |\hat{\theta}_i|\right)} \nonumber \\
 &= -\ell_N(\hat{\boldsymbol{\theta}}) + k \log n + \sum_j {\log c_j} - \frac{d}{2} \log 3 \nonumber \\ &+ \sum_{i=1}^d{\left(\frac{1}{2}\log I_{ii} + \log |\hat{\theta}_i|\right)}. \nonumber
\end{align*}
\endgroup

The proposed DL was tasked with ranking a set of enumerated equations.
The results showed that, while MDL principle does not guarantee the choice of the optimal solution to be the first in rank, it is still within the top choices.
The MDL principle provides a selection of alternative models that can be investigated more carefully to assert the correct hypothesis.

\section{Discussion}

As the aftermath of this survey, we noticed a growing interest in extending UQ for the particularities of SR, and some points of the current state must be highlighted.
Even though traditional UQ methods for regression analysis are well adopted in practical applications and can be easily integrated into SR frameworks, only a few of the surveyed papers~\cite{zhao_uncertainty_2025,bomarito_automated_2023,thuong_combining_2017,de_franca_prediction_2022} actually calculate confidence, credible, or prediction intervals. 
From the $18$ surveyed papers, only $3$ were published before $2020$, showing that only recently the field started to acknowledge the importance of this topic.
Despite their recency, the papers tackle different aspects of UQ specific to SR, such as establishing priors over model structures, selecting the most likely model, or approximating the posterior distribution when the model structure is not specified \emph{a priori}, as in traditional regression analysis.
Only a minority of relevant papers cover \textbf{frequentist approaches}, as estimating the aleatoric uncertainty can be challenging in frequentist settings: SR models are flexible and can easily under- or overfit, and can thus not serve as an unbiased estimator, which is however required to estimate $\sigma^2$.
A reasonable non-Bayesian alternative is to generate the model with random components~\cite{schmidt_learning_2007}, which includes the prediction intervals \emph{for free}, but makes the search more difficult as the algorithm must learn the correct structure together with the noise distribution.

The remainder of relevant papers in our survey used \textbf{Bayesian approaches}, which inherently enable UQ through applying sampling-based methods, such as MCMC and SMC.
One line of research focuses on the challenges of integrating prior knowledge about the model structure into the Bayesian model priors~\cite{guimera_bayesian_2020,guimera_bayesian_2026,bartlett_priors_2023}. 
Since Bayesian posteriors are typically a collection of model structures, more robust predictions can be made using a weighted average of the top-$K$ models weighted by their likelihood~\cite{agapitos_ensemble_2014}.
Although often neglected, the measurement errors of the independent variable can be integrated into the search by means of an adapted likelihood function~\cite{bartlett_marginalised_2023}, with added computational costs.

Another important application of UQ in SR is model selection, where uncertainties must be taken into consideration during model evaluation to assess how much the likelihood function can be trusted or if the search algorithm should propose less complex model to avoid overfitting~\cite{nikolaev_regularization_2001,bartlett_exhaustive_2024}.

In our survey, we covered the research streams of Bayesian SR with SMC~\cite{bomarito_automated_2023,bomarito_bayesian_2022,bomarito_bayesian_2026,leser_comparing_2024} and Bayesian forest of probabilistic trees~\cite{roy_hierarchical_2025,roy_vasst_2026} in more detail, as they introduced novelty into UQ for SR.
Both aim to develop posterior distributions over model structures, but in fundamentally different ways: \citet{bomarito_bayesian_2026} directly approximate the posterior as a weighted set of discrete trees, similar to the final population of a GP algorithm, and sampling from the posterior returns models from this set.
The primary latent variable in the inference process is the model structure, and potentially non-linear parameters are treated as part of the model structure and optimized separately.
This hinders closed-form estimation of the marginal likelihood and motivates the use of SMC. 
To the contrary, \citet{roy_vasst_2026} develop a posterior over the tree generating process, with parametric feature and operator distributions at every node.
Every draw from the posterior samples a new hard tree from the soft tree skeleton.
The conjugacy between linear parameters and likelihood enables the analytical marginalization over the parameter space, but limits the models to be linear in their parameters. 
Moreover, the model structure is defined by multiple latent variables, and optimized using a mean-field variational inference approach, which ignores dependencies between nodes in a tree. 

The SMC-based approach with multiple particles exploring the posterior space in~\cite{bomarito_bayesian_2026} can generally capture multimodal posteriors, if the set of particles is large and diverse enough. 
The posterior is also set up in a way that models with a high posterior weight are sampled more often.
We assume that the per-node feature and operator distributions can, in principle, capture multimodality locally in~\cite{roy_vasst_2026} (e.g., ``$+$'' and ``$\times$'' have the same probability at the root node), but the mean-field approach ignores node dependencies. 
As a result, optimal decisions are only made locally on a node level, but the combination of many locally optimal decisions does not guarantee an optimal overall model structure, especially in multi-modal landscapes.
Therefore, we question whether repeated sampling from soft posterior trees approximates the true model structure posterior closely.

The results of both papers raise the broader question of how to verify whether the approximated posterior is close to the true one, and how to ultimately quantify model structure uncertainty (approximation uncertainty).
During the survey, we identified a major challenge when integrating a prior over model structures into the definition of the posterior.
Ignoring dependencies between nodes in the prior leads to locally optimal decisions that may translate to a non-optimal model when sampling from the posterior.
On the other hand, considering this dependency requires a large dataset of equations, if learning the prior before applying SR as in~\cite{guimera_bayesian_2020,bartlett_priors_2023}, or a large number of tree samples, if learning the parameters of the priors during inference as in~\cite{roy_vasst_2026}. 

\section{Conclusion}
This work discussed uncertainty quantification~(UQ) for symbolic regression~(SR), with the goal of providing a solid theoretical foundation and surveying relevant literature.
We introduced the main frequentist and Bayesian UQ methods, such as confidence and credible intervals, together with concepts that enable UQ in regression, such as likelihood and Fisher information.
 
The overall small number of relevant articles allowed us to survey emerging approaches in greater detail, specifically Bayesian formulations of SR.
Despite their large potential, especially since UQ comes almost for free in sampling-based approaches, we also illustrated their current limitations.
Overall, we conclude that UQ in SR remains largely underexplored, and its importance is systematically underestimated.
Therefore, we emphasize that future works should adopt established UQ methods for SR, along with a careful analysis of the data and underlying noise, as this understanding is essential for developing reliable SR models.

\footnotesize{
\bibliography{references,control}}

\newpage
\normalsize
\appendices

\section{Supplementary Material for basic concepts of UQ}

\subsection{Likelihood and Fisher information}
\label{sec:appendix_likelihood}

As introduced in Sec.~\ref{sec:likelihood_fisher}
, the Jacobian is a matrix of partial derivatives of the model outputs with respect to the parameters, and has the shape $N \times d$, corresponding to $N$ data points and $d$ parameters in the equation. 
This matrix can be written as:
\begin{align*}
    J(\boldsymbol{\theta})
    =
    \begin{pmatrix}
    \frac{\partial f(\mathbf{x}_1;\boldsymbol{\theta})}{\partial \theta_1}
    &
    \frac{\partial f(\mathbf{x}_1;\boldsymbol{\theta})}{\partial \theta_2}
    &
    \cdots
    &
    \frac{\partial f(\mathbf{x}_1;\boldsymbol{\theta})}{\partial \theta_d}
    \\[8pt]
    \frac{\partial f(\mathbf{x}_2;\boldsymbol{\theta})}{\partial \theta_1}
    &
    \frac{\partial f(\mathbf{x}_2;\boldsymbol{\theta})}{\partial \theta_2}
    &
    \cdots
    &
    \frac{\partial f(\mathbf{x}_2;\boldsymbol{\theta})}{\partial \theta_d}
    \\[8pt]
    \vdots & \vdots & \ddots & \vdots
    \\[8pt]
    \frac{\partial f(\mathbf{x}_N;\boldsymbol{\theta})}{\partial \theta_1}
    &
    \frac{\partial f(\mathbf{x}_N;\boldsymbol{\theta})}{\partial \theta_2}
    &
    \cdots
    &
    \frac{\partial f(\mathbf{x}_N;\boldsymbol{\theta})}{\partial \theta_d}
    \end{pmatrix} 
    \in \mathbb{R}^{N \times d}.
\end{align*}

When optimizing the negative log-likelihood, the first-order optimality condition is
\begin{align}
    J(\hat{\boldsymbol{\theta}})^\top \bigl(y - f(\hat{\boldsymbol{\theta}})\bigr) = 0,
\end{align}
and this system of nonlinear equations is solved typically with numerical methods, such as Gauss-Newton or Levenberg-Marquardt. 

In Sec.~\ref{sec:likelihood_fisher}
, we also introduced the observed Fisher information as the negative Hessian of the log-likelihood.
While reviewing the corresponding literature, we noticed confusion between the \textit{observed} and \textit{expected} information, which we aim to resolve here. 
The expected Fisher information is defined as the expected value of the observed information over all possible observations, calculated as:
\begin{align*}
    \mathcal{I}(\boldsymbol{\theta}) &= \mathbb{E_\theta}[I(\boldsymbol{\theta})] 
    =\mathbb{E}[\nabla_{\boldsymbol{\theta}} \ell_N(\boldsymbol{\theta})^\top \nabla_{\boldsymbol{\theta}} \ell_N(\boldsymbol{\theta})] \\
    &=\int_x{I(\boldsymbol{\theta})L_N(\boldsymbol{\theta})dx}.
\end{align*}

Thus, the expected Fisher information requires knowledge of the underlying probability model to compute the expectation over the data-generating distribution.
However, the underlying data generating process is unknown in practice, and the expectation is typically analytically intractable, as it involves integration over all possible observations, see for instance~\cite{pawitan_all_2001}.
Therefore, this work focuses on the observed Fisher information $I(\boldsymbol{\theta})$ rather than the expected Fisher information $\mathcal{I}(\boldsymbol{\theta})$.
This is in line with the reviewed publications on symbolic regression that make use of Fisher information, specifically~\cite{bartlett_exhaustive_2024,bartlett_priors_2023}.

\subsection{Hypothesis testing}
\label{sec:appendix_hypothesis}

A dual to the concept of CI (Sec.~\ref{sec:ci-pi}) 
is the hypothesis test that verifies whether if a certain value is plausible, given the calculated intervals, leading to a \emph{reject} or \emph{fail to reject} decision.
In a parametric model, parameters appear as multiplicative or additive components in a sub-expression.
For example, in the expression $\sin(\theta_0 e^x + \theta_1)$, $\theta_0$ multiplies $g(x) = e^x$ and $\theta_1$ is added to this same sub-expression.
Similarly, the expression $e^{\theta_0}x$ would be equivalent to $\theta_0 x$, since we fit the parameter to the data.

A question that arises during regression analysis is whether a certain parameter $\theta_i = 0$.
In practice, this would mean that the corresponding sub-expression is unnecessary to the model and, thus, can be simplified.
For this purpose we perform a \textbf{hypothesis test}, in which we pose a null and an alternative hypothesis:
\begin{align*}
    \mathbb{H}_0&: \theta_i = 0, \\
    \mathbb{H}_\alpha&: \theta_i \neq 0.
\end{align*}

To test this hypothesis, we can use the \textbf{Wald test} and calculate the t-statistic with:
\begin{align*}
    t = \frac{\theta_i - 0}{\operatorname{s}_{\theta_i}}.
\end{align*}

Given this value and the desired $\alpha$, we obtain a $p$-value from a statistic table of a Student-t distribution and reject the hypothesis if $p < \alpha$~\cite[Sec.~5.5.3]{murphy_probabilistic_2022}.
When $p = 0.05$, there is a $5\%$ likelihood that the observed value of $\theta_i$ arose from random sampling error rather than a true underlying effect.

\subsection{Model selection methods}
\label{sec:appendix_modelselection}
While Sec.~\ref{sec:model_selection} 
of this paper focuses on advanced methods of model selection for SR, we provide the details for cross validation, AIC and BIC subsequently for completeness.

\subsubsection{Cross validation}

A straightforward approach to detect overfitting caused by aleatoric uncertainty is \textbf{cross validation}, where the data is randomly split into different sets for training and validation.
By fitting the model into the training set and evaluating it into the validation, we get a more reliable unbiased estimation of the likelihood.
The idea is that, if the model overfits the training data, it will perform poorly on the validation set.
One drawback of this approach is its sensitivity to the sampling of the splits, potentially creating two datasets with different distributions~\cite[Sec.~4.5.5]{murphy_probabilistic_2022}. 

A better approach that avoids this pitfall is the $k$-fold cross validation, in which the training data is randomly split into $k$ equally sized sets. 
At every iteration, $k-1$ sets are used as the training data and the remaining one as the validation set.
Afterwards, the scores are averaged over the folds, giving an unbiased estimation of the quality of the model.
 
\subsubsection{Akaike Information Criterion} 

The \textbf{Akaike Information Criterion}~(AIC) penalizes the model linearly by the number of parameters $d$~\cite[Sec.~3.5]{pawitan_all_2001}. 
The codelength increases by $2$ for every additional unit of the log-likelihood or for every additional parameter
\begin{align*}
    \operatorname{AIC} = -2 \ell_N(\hat{\boldsymbol{\theta}}) + 2d.
\end{align*}

In this sense, this criterion emphasizes the minimization of the negative log-likelihood, with a small penalty to complexity.
The number of samples $N$ is implicitly considered in the standard AIC, as $\ell_N$ is an additive term of the error over the number of samples (see Eq.~4). 
Considering two hypothesis $h_1$ and $h_2$, where $h_2$ has exactly one additional parameter than $h_1$, then $h_2$ is preferred over $h_1$ if $\ell_N(h_2) - \ell_N(h_1) > 1$.
Thus, the decrease of the log-likelihood due to an additional parameter is given by $N \mathbb{E}[\frac{\left(y_n - f(x_n, \boldsymbol{\theta})\right)^2}{2 \sigma^2}]$.
For example, when an additional parameter causes an improvement of the expected scaled per-sample-error by $0.002$, with $1\,000$ samples the difference between the log-likelihoods would be $2>1$, and thus the more complex model would be selected.
This can be alleviated with the corrected AIC:
\begin{align*}
    \operatorname{AIC}_c = -2 \ell_N(\hat{\boldsymbol{\theta}}) + \frac{2d(d+1)}{N - d -1}.
\end{align*}

\subsubsection{Bayesian Information Criterion} \label{sec:bic}

The \textbf{Bayesian Information Criterion}~(BIC) penalizes the model by its number of parameters with a magnitude proportional to the number of samples
\begin{align*}
    \operatorname{BIC} = -2 \ell_N(\hat{\boldsymbol{\theta}}) + d\log(N).
\end{align*}

The complexity penalty is scaled by the logarithm of $N$, thus under a large sample regime, models with fewer parameters are preferred~\cite[Sec.~5.2.5.1]{murphy_probabilistic_2022}.
Compared to the classical AIC, the BIC is considered a \textbf{consistent model selection technique}, as it becomes increasingly skeptical about the addition of potentially unnecessary parameters with increasing sample size. 
Thus, it is more oriented towards selecting a ``true'' underlying model. 

Summing up the model selection methods presented here and in Sec.~\ref{sec:model_selection}
, due to uncertainties associated with the regression task, the models should not be expected to fit the data exactly.
As such, there must be criteria to select the best model based on the balance between accuracy and complexity.
Next to cross-validation as an established method to detect overfitting, we introduced AIC, BIC, and FBF and the MDL principle.
In short, they differ in the questions they answer: AIC is designed to answer how well a model predicts new data. 
BIC, on the other hand, aims to identify the most plausible model in relation to the size of the dataset, assuming that the true model is in the candidate set. 
FBF compares to what extent two models support the observed data, including a correction for improper priors. 
And finally, the MDL principle seeks the model that achieves the greatest compression of the data composed of an approximation error and a complexity term.

\end{document}